%% file: sum_expressivity_pub.tex
\documentclass{article}

\pdfoutput=1

\usepackage{sum_expressivity_style}
\usepackage{times}
\usepackage{soul}
\usepackage{url}
\usepackage[hidelinks]{hyperref}
\usepackage[utf8]{inputenc}
\usepackage[small]{caption}
\usepackage{graphicx}
\usepackage{booktabs}
\usepackage{algorithm}
\usepackage{algorithmic}
\usepackage[switch]{lineno}

\usepackage{caption}
\usepackage{amsmath,amsfonts, mathdots, amssymb, latexsym, amsbsy,bm,amsthm,mathtools}
\usepackage{hyperref}
\usepackage{braket}
\usepackage{xcolor}
\usepackage{comment}

\usepackage{xspace}
\usepackage{times}
\usepackage{framed}
\usepackage{subcaption}
\usepackage[nameinlink]{cleveref}

\usepackage{txfonts}

\newcommand{\ceil}[1]{\left \lceil #1 \right \rceil }
\newcommand{\abs}[1]{\left\lvert #1 \right\rvert}
\newcommand{\norm}[1]{\left\lVert #1 \right\rVert}

\newtheorem{thm}{Theorem}[section]
\newtheorem{theorem}[thm]{Theorem}
\newtheorem{lemma}[thm]{Lemma}

\newtheorem{corollary}[thm]{Corollary}
\theoremstyle{definition}

\renewcommand{\phi}{\varphi}
\renewcommand{\epsilon}{\varepsilon}

\newcommand{\Nat}{{\mathbb N}}
\newcommand{\PNat}{{\mathbb N}_{>0}}
\newcommand{\Real}{{\mathbb R}}

\newcommand{\Rat}{{\mathbb Q}}

\newcommand{\CA}{{\mathcal A}}
\newcommand{\CB}{{\mathcal B}}

\newcommand{\CG}{{\mathcal G}}

\newcommand{\CL}{{\mathcal L}}

\newcommand{\CN}{{\mathcal N}}

\newcommand{\CZ}{{\mathcal Z}}

\newcommand{\FF}{{\mathfrak F}}

\newcommand{\Fa}{{\mathfrak a}}

\newcommand{\Fo}{{\mathfrak o}}
\newcommand{\Fr}{{\mathfrak r}}
\newcommand{\Fro}{{\Fr\Fo}}

\DeclareMathOperator{\depth}{depth}

\usepackage[auth-lg]{authblk}
\newcommand{\mean}{\operatorname{avg}}
\newcommand{\lsum}{\operatorname{sum}}
\newcommand{\sag}{$\operatorname{Sum}$-GNN }
\newcommand{\aag}{$\operatorname{Mean}$-GNN }
\newcommand{\mxag}{$\operatorname{Max}$-GNN }
\newcommand{\mupa}{$\operatorname{MUPA}$-GNN }

\newcommand{\sags}{$\operatorname{Sum}$-GNNs }
\newcommand{\aags}{$\operatorname{Mean}$-GNNs }
\newcommand{\mxags}{$\operatorname{Max}$-GNNs }

\newcommand{\sagns}{$\operatorname{Sum}$-GNN}

\newcommand{\mxagns}{$\operatorname{Max}$-GNN}

\newcommand{\sagsns}{$\operatorname{Sum}$-GNNs}
\newcommand{\aagsns}{$\operatorname{Mean}$-GNNs}
\newcommand{\mxagsns}{$\operatorname{Max}$-GNNs}
\newcommand{\mupasns}{$\operatorname{MUPA}$-GNNs}

\newcommand{\subsumesp}[1]{\geq^{_{#1}}}
\newcommand{\subsume}{\geq}

\newcommand{\nsubsumesp}[1]{\not\geq^{_{#1}}}
\newcommand{\norme}[1]{\norm{#1}_2}
\newcommand{\normi}[1]{\norm{#1}_I}

\usepackage[multiple]{footmisc}
\usepackage{pgfplots}
\usetikzlibrary{math}
\begin{document}

\input{sum_expressivity_body.tex}

\input{sum_expressivity_experiments.tex}

\cleardoublepage

\bibliographystyle{sum_expressivity_literature_style_named}
\bibliography{sum_expressivity_literature}

\cleardoublepage
\input{sum_expressivity_appendix}
\input{sum_expressivity_experiments_appendix.tex}

\end{document}

%% file: sum_expressivity_body.tex
\title{Some Might Say All You Need Is Sum}
\author{
Eran Rosenbluth$^{1,}$\thanks{Funded by the German Research Council (DFG), RTG 2236 (UnRAVeL)}\and
Jan Toenshoff$^{1,}$\thanks{Funded by the German Research Council (DFG), grants GR 1492/16-1; KI 2348/1-1 “Quantitative Reasoning About Database Queries”}
\and
Martin Grohe$^{1}$\\
[rosenbluth $\mid$ toenshoff $\mid$ grohe]@informatik.rwth-aachen.de\\
$^1${RWTH Aachen University}
}
\maketitle

\begin{abstract}
The expressivity of Graph Neural Networks (GNNs) is dependent on the aggregation functions they employ. Theoretical works have pointed towards Sum aggregation GNNs subsuming every other GNNs, while certain practical works have observed a clear advantage to using Mean and Max. An examination of the theoretical guarantee identifies two caveats. First, it is size-restricted, that is, the power of every specific GNN is limited to graphs of a specific size. Successfully processing larger graphs may require an other GNN, and so on. Second, it concerns the power to distinguish non-isomorphic graphs, not the power to approximate general functions on graphs, and the former does not necessarily imply the latter.

It is desired that a GNN's usability will not be limited to graphs of any specific size.
Therefore, we explore the realm of unrestricted-size expressivity. We prove that basic functions, which can be computed exactly by Mean or Max GNNs, are inapproximable by any Sum GNN. We prove that under certain restrictions, every Mean or Max GNN can be approximated by a Sum GNN, but even there, a combination of (Sum, [Mean/Max]) is more expressive than Sum alone. Lastly, we prove further expressivity limitations for GNNs with a broad class of aggregations.
\end{abstract}

\section{Introduction}\label{auto_label_1}
Message passing graph neural networks (GNNs) are a fundamental deep learning architecture for machine learning on graphs. Most state-of-the-art machine learning techniques for graphs are based on GNNs. It is therefore worthwhile to understand their theoretical properties. Expressivity is one important aspect: which functions on graphs or their vertices can be computed by GNN models? To start with, functions computed by GNNs are always isomorphism invariant, or equivariant for node-level functions.
A second important feature of GNNs is that a GNN can operate on input graphs of every size, since it is defined as a series of node-level computations with an optional graph-aggregating readout computation. These are desirable features that motivated the introduction of GNNs in the first place and may be seen as a crucial factor for their success. Research on the expressivity of GNNs has had a considerable impact in the field.

A GNN computation transforms a graph with an initial \emph{feature map} (a.k.a.\ \emph{graph signal} or \emph{node embedding}) into a new feature map. The new map can represent a node-level function or can be ``read out'' as a function of the whole graph. The computation is carried out by a finite sequence of separate \emph{layers}. On each layer, each node sends a real-valued message vector which depends on its current feature vector, to all its neighbours. Then each node aggregates the messages it receives, using an order-invariant multiset function, typically being entrywise summation (Sum), mean (Mean), or maximum (Max). Finally, the node features are updated using a neural network which receives as arguments the aggregation value and the node's current feature. In the eyes of a GNN all vertices are euqal: the message, aggregation and update functions of every layer are identical for every node, making GNNs auto-scalable and isomorphism-invariant.

By now, numerous works have researched the expressivity of GNNs considering various variants of them. However, many of the theoretical results have the following caveats:
\\1. The expressivity considered is \emph{non-uniform}: for a function that is defined on graphs of all sizes, it is
asked if for every $n$ there exists a GNN that expresses the function on graphs of size $n$.
The expressing GNN may depend on $n$, and it may even be exponentially large in $n$. For some proofs, this exponential blow-up is necessary \cite{abboud2020surprising,xu2018powerful}.
This notion of expressivity is in contrast to \emph{uniform} expressivity: for a function that is defined on graphs of all sizes, asking whether there exists one GNN that expresses the function on graphs of all sizes.
In addition to being a significantly weaker theoretical notion, non-uniform expressivity leaves much to be desired also from a practical standpoint: It implies that a GNN may be no good for graphs of sizes larger than the sizes well-represented in the training data. This means that training may have to be done on very large graphs, and may have to be often repeated.
\\2. The expressivity considered is the power to distinguish non-isomorphic graphs. 
A key theoretical result is the characterisation of the power of GNNs in terms of the Weisfeiler-Leman (WL) isomorphism test \cite{morris2019weisfeiler,xu2018powerful}, and subsequent works have used WL as a yardstick (see 'Related Work'). In applications of GNNs though, the goal is not to distinguish graphs but to regress or classify them or their nodes. There seem to be a hidden assumption that higher distinguishing power implies better ability to express general functions. While this is indeed the case in some settings \cite{chen2019equivalence}, it is not the case with uniform expressivity notion.

Our goal is to better understand the role that the aggregation function plays in the expressivity of GNNs. Specifically, we ask:
Do Sum aggregation GNNs subsume Mean and Max GNNs, in terms of uniform expressivity of general functions?
\\A common perception is that an answer is already found in ~\cite{xu2018powerful}: \sags strictly subsume all other aggregations GNNs. Examining the details though, what is actually proven there is: in the non-uniform notion, considering a finite input domain, the distinguishing power of \sags subsume the distinguishing power of all other aggregations GNNs.
Furthermore, in practice it has been observed that for certain tasks there is a clear advantage to using Mean and Max aggregations \cite{cappart2021combinatorial,hamilton2017inductive,tonshoff2022one}, with one of the most common models in practice using a variation of Mean aggregation \cite{kipf2016semi}. While the difference between theoretical belief and practical evidence may be attributed to a learnability rather than to expressivity, it calls for better theoretical understanding of expressivity.

\subsection{Our Contribution}
All our results are in the uniform expressivity notion. Mainly, we prove that \sags do not subsume \aags nor \mxags (and vice versa), in terms of vertices-embedding expressivity as well as graph-embedding expressivity. The statements in this paper consider additive approximation, yet the no-subsumption ones hold true also for multiplicative approximation.

\begin{itemize}
\item Advantage Sum. For the sake of completeness, 
in \Cref{sec:sums_place} we prove that even with single-value input features, the neighbors-sum function which can be trivially exactly computed by a \sag cannot be approximated by any \aag or \mxagns.

\item Sum subsumes. In \Cref{sec:sum_enough} we prove that if the input features are bounded, \sags can approximate all \aags or \mxagsns, though not without an increase in size which depends polynomially on the required accuracy, and exponentially on the depth of the approximated \aags or \mxagsns.

\item Advantage Mean and Max. In \Cref{subsec:unbounded} we show that if we allow unbounded input features then functions that are exactly computable by \aags; \mxagsns; and others, cannot be approximated by \sagsns.

\item Essential also with finite input-features domain. In \Cref{subsec:simple} we prove that even with just single-value input features, there are functions that can be exactly computed by a (Sum, Mean)-GNN (a GNN that use both Sum-aggregation and Mean-aggregation) or by a (Sum, Max)-GNN, but cannot be approximated by \sagsns. 

\item
The world is not enough. In \Cref{sec:more_sum}, we examine GNNs with any finite combination of Sum; Mean; Max and other aggregations, and prove upper bounds on their expressivity already in the single-value input features setting.
\end{itemize}

Lastly, in \Cref{sec:experiment} we experiment with synthetic data and observe that what we proved to be expressible is to an extent also learnable, and that in practice inexpressivity is manifested in a significantly higher error than implied in theory.

All proofs, some of the lemmas, and extended illustration and analysis of the experimentation, are found in the appendix.

\subsection{Related Work}
The term Graph Neural Network, along with one of the basic models of GNNs, was introduced in \cite{scarselli2008graph}.
Since then, more than a few works have explored aspects of expressivity of GNNs. Some have explored the distinguishing power of different models of GNNs \cite{abboud2020surprising,barcelo2021graph,geerts2022expressiveness,maron2019provably,morris2019weisfeiler,morris2020weisfeiler,sato2021random}, and some have examined the expressivity of GNNs depending on the aggregations they use \cite{corso2020principal,xu2018powerful}. In \cite{chen2019equivalence}, a connection between distinguishing power and function approximation is described. In all of the above, the non-uniform notion was considered. In the uniform notion, it was proven that \sags can express every logical formula in Guarded Countable Logic with 2 variables (GC2) \cite{barcelo2020expressive,barcelo2020logical}. A theoretical survey of the expressivity of GNNs is found in \cite{grohe2021logic}, and a practical survey of different models of GNNs is found in \cite{wu2020comprehensive}.

\section{Preliminaries}\label{auto_label_2}
By $\Nat,\PNat,\Rat,\Real$ we denote the sets of nonnegative integers,
positive integers, rational numbers, an d real numbers, respectively. 
For $a,b\in\Nat:a\leq b$ we denote the set $\{n\in\Nat : a\leq n\leq b\}$ by $[a..b]$.
For $b\in\PNat$ we denote the set $[1..b]$ by $[b]$. 
For $a,b\in\Real:a\leq b$, we denote the set $\{r\in\Real : a\leq r\leq b\}$ by $[a,b]$ .
We may use the terms "average" and "mean" interchangeably to denote the arithmetic mean.
We use "\{\}" as notation for a multiset.
Let $x\in\Real,b\in\PNat$, we define ${\{x\} \choose b}\coloneqq\{x,\ldots,x\}$ the multiset consisting of $b$ instances of $x$.
Let $d\in\PNat$ and let a vector $v\in\Real^d$, we define $\abs{v}\coloneqq max(\abs{v_i}_{i\in[d]})$.
Let two vectors $u,v\in\Real^d$, we define $'\leq'$: $u\leq v \Leftrightarrow \forall i\in[d] u_i\leq v_i$.

\subsection{Graphs}
\label{sec:graph}
An \emph{undirected graph} $G=\langle V(G),E(G)\rangle$ is a pair, $V(G)$ being a set of vertices and $E(G)\subseteq\{\{u,v\}\mid u,v\in V(G)\}$ being a set of undirected edges.
For a vertex $v\in V(G)$ we denote by $N(v)\coloneqq\{w\in V(G)\mid \{w, v\}\in E(G)\}$ the neighbourhood of $v$ in $G$, and we denote the size of it by $n_v\coloneqq|N(v)|$.

A (vertex) \emph{featured graph} $G=\langle V(G),E(G),S^d,Z(G)\rangle$ is a $4$-tuple
being a graph with a \emph{feature map} $Z(G):V(G)\rightarrow S^d$, mapping each vertex to a $d$-tuple over a set $S$. We denote the set of graphs featured over $S^d$ by $\CG_{S^d}$, we define ${\CG_S\coloneqq\bigcup_{d\in \Nat}\CG_{S^d}}$, and we denote the set of all featured graphs by $\CG_*$. The special set of graphs featured over \{1\} is denoted $\CG_1$.
We denote the set of all feature maps that map to $S^d$ by $\CZ_{S^d}$, we denote $\bigcup_{d\in \Nat}\CZ_{S^d}$ by $\CZ_S$, and we denote the set of all feature maps by $\CZ_*$. Let a featured-graph domain $D\subseteq\CG_*$, a mapping $f:\CG_{D}\rightarrow \CZ_*$ to new feature maps is called a \emph{feature transformation}.

For a featured graph $G$ and a vertex $v\in V(G)$ we define $\lsum(v)\coloneqq \Sigma_{w\in N(v)}Z(G)(w)$, $\mean(v)\coloneqq \frac{1}{n_v}\lsum(v)$, and $\max(v)\coloneqq\max(Z(G)(w):w\in N(v))$.
In this paper, we consider the size of a graph $G$ to be its number of vertices, that is, $\abs{G}\coloneqq\abs{V(G)}$.

\subsection{Feedforward Neural Networks}
A \emph{feedforward neural network (FNN)} $\FF$ is directed acyclic graph
where each edge $e$ carries a \emph{weight} $w_e^\FF \in\Real$, each
node $v$ of positive in-degree carries a \emph{bias} $b_v^\FF \in\Real$, and each node $v$
has an associated continuous \emph{activation function}
$\Fa_v^\FF:\Real\to\Real$. The nodes of in-degree $0$,
usually $X_1,\ldots,X_p$, are the \emph{input nodes} and the nodes of
out-degree $0$, usually $Y_1,\ldots,Y_q$, are the \emph{output
  nodes}. We denote the underlying directed graph of an FNN $\FF$ by
$(V(\FF),E(\FF))$, and we call
$\big(V(\FF),E(\FF),(\Fa^\FF_v)_{v\in V(\FF)}\big)$ the
\emph{architecture of $\FF$}, notated $A(\FF)$. We drop the indices ${}^\FF$ at the weights
and the activation function if $\FF$ is clear from the context.

The \emph{input dimension}
of an FNN is the number of input nodes, and the \emph{output
  dimension} is the number of output nodes. The \emph{depth}
$\depth(\FF)$ of an FNN $\FF$ is the maximum length of a path from an input node to an
output node.

To define the semantics, let $\FF$ be an FNN of input dimension $p$ and
output dimension $q$. For each node $v\in V(\FF)$, we define a function
$f_{\FF,v}:\Real^p\to\Real$ by $f_{\FF,X_i}(x_1,\ldots,x_p)\coloneqq x_i$
for the $i$th input node $X_i$ and 
\[
  f_{\FF,v}(\vec x)\coloneqq
  \Fa_v\left(b_v+\sum_{j=1}^kf_{\FF,u_j}(\vec x)\cdot w_{e_j}\right)
\]
for every node $v$ with incoming edges $e_j=(u_j,v)$. Then $\FF$ computes
the function $f_\FF:\Real^p\to\Real^q$ defined by
\[
  f_\FF(\vec x)\coloneqq\big(f_{\FF,Y_1}(\vec x),\ldots, f_{\FF,Y_q}(\vec
  x)\big)
\]

Let $\FF$ an FNN, we consider the size of $\FF$ to be the size of its underlying graph. That is, $\abs{\FF}=\abs{V(\FF)}$.

A common activation function is the ReLU activation, defined as $ReLU(x)\coloneqq max(0,x)$. In this paper, we assume all FNNs to be ReLU activated. ReLU activated FNNs subsume every finitely-many-pieces piecewise-linear activated FNN, thus the results of this paper hold true for every such FNNs.
Every ReLU activated FNN $\FF$ is Lipschitz-Continuous. That is, there exists a minimal $a_\FF\in\Real_{\geq 0}$ such that for every input and output coordinates $(i,j)$, for every specific input arguments $x_1,\ldots,x_n$, and for every $\delta>0$, it holds that \[\abs{f_\FF(x_1,\ldots,x_n)_j-f_\FF(x_1,\ldots x_{i-1},x_i+\delta,\ldots,x_n)_j}/\delta\leq a_\FF\]
We call $a_\FF$ the \emph{Lipschitz-Constant} of $f$.

\subsection{Graph Neural Networks}
Several GNN models are described in the literature. In this paper, we define and consider the Aggregate-Combine (AC-GNN) model \cite{xu2018powerful,barcelo2020expressive}. Some of our results extend straightforwardly to the messaging scheme of MPNN \cite{gilmer2017neural}, yet such extensions are out of scope of this paper.

A \emph{GNN layer}, of input and output (I/O) dimensions ${p;q}$, is a pair $(\FF,agg)$ such that: $\FF$ is an FNN of I/O dimensions ${2p;q}$, and $agg$ is an order-invariant $p$-dimension multiset-to-one aggregation function.
An $m$-layer GNN $\CN=((\FF_1,agg_1),\ldots$ $,(\FF_m,agg_m))$, of I/O dimensions $p;q$, is a sequence of $m$ GNN layers of I/O dimensions $p^{(i)};q^{(i)}$ such that: $p^{(1)}=p$, $q^{(m)}=q$ and $\forall i\in[m-1]\ p^{(i+1)}=q^{(i)}$. It determines a series of $m$ feature transformations as follows: Let a graph $G\in\CG_{\Real^{p}}$ and vertex $v\in V(G)$, then $\CN^{(0)}(G,v)\coloneqq Z(G)(v)$, and for ${i\in[m]}$ we define a transformation
$$\CN^{(i)}(G,v)\coloneqq f_{\FF_i}(\CN^{(i-1)}(G,v),agg_i(\CN^{(i-1)}(G,w):{w\in N(v)}))$$

We notate by $\CN(G,v)\coloneqq \CN^{(m)}(G,v)$ the final output of $\CN$ for $v$. We define the size of $\CN$ to be ${\abs{\CN}\coloneqq\Sigma_{i\in[m]}\abs{\FF_i}}$ the sum of its underlying FNNs' sizes. We call $\big((A(\FF_1),agg_1),\ldots,(A(\FF_m),agg_m)\big)$ the \emph{architecture} of $\CN$, notated $A(\CN)$, and say that $\CN$ \emph{realizes} $A(\CN)$. For an aggregation function $agg$, we denote by $agg$-GNNs the class of GNNs for which $\forall i\in[m]\ agg_i=agg$. For aggregation functions $agg_1,agg_2$, we denote by $(agg_1,agg_2)$-GNNs the class of GNNs with $m=2n$ layers such that ${\forall i\in[n]\ agg_{2i-1}=agg_1, agg_{2i}=agg_2}$.

\subsection{Expressivity}
Let $p,q\in\Nat$, and a set $S$. Let $F=\{f:\CG_{S^p}\rightarrow \CZ_{\Real^q}\}$ a set of feature transformations, and let a feature transformation ${h:\CG_{S^p}\rightarrow \CZ_{\Real^q}}$. We say $F$ \emph{uniformly additively approximates} $h$, notated $F\approx h$, if and only if
$$\forall \epsilon>0 \exists f\in F: \forall G\in\CG_{S^p} \forall v\in V(G)\ \abs{f(G)(v)-h(G)(v)}\leq\epsilon$$
The essence of uniformity is that one function "works" for graphs of all sizes, unlike non-uniformity where it is enough to have a specific function for each specific size of input graphs. The proximity measure is additive - as opposed to multiplicative where it is required that ${\abs{\frac{f(G)(v)-h(G)(v)}{h(G)(v)}}\leq\epsilon}$. In this paper, approximation always means uniform additive approximation and we use the term "approximates" synonymously with \emph{expresses}. Although our no-approximation statements consider additive approximation, they hold true also for multiplicative approximation, and the respective proofs (in the appendix) require not much additional argumentation to show that.

Let $F,H$ be sets of feature transformations $f:\CG_{S^p}\rightarrow \CZ_{\Real^q}$, we say $F$ \emph{subsumes} $H$, notated $F\subsume H$ if and only if for every $h:\CG_{S^p}\rightarrow \CZ_{\Real^q}$ it holds that $H\approx h\Rightarrow F\approx h$.
If the subsumption holds only for graphs featured with a subset $T^p\subset S^p$ we notate it as $F\subsumesp{T} H$.

Let $p,q\in\Nat$. We call an order-invariant mapping ${f:\CZ_{\Real^p}\rightarrow \Real^q}$, from feature maps to $q$-tuples, a \emph{readout function}. Both $\lsum$ and $\mean$ are commonly used to aggregate feature maps, possibly followed by an FNN that maps the aggregation value to a final output.
We call a mapping ${f:\CG_{S^p}\rightarrow \Real^q}$, from featured graphs to $q$-tuples, a \emph{graph embedding}. 
Let $w\in\Nat$, let a set of feature transformations $F={\{f:\CG_{S^p}\rightarrow \CZ_{\Real^q}\}}$, and let a readout ${r:\CZ_{\Real^q}\rightarrow \Real^w}$, we notate the set of embeddings ${\{r\circ f : f\in F\}}$ by $r\circ F$.
We use the expressivity terms and notations defined for feature transformations, for graph embeddings as well.

\section{Mean and Max Do Not Subsume}\label{sec:sums_place}
\begin{figure*}[h]
\begin{minipage}{.32\linewidth}
    \begin{center}
            \includegraphics[width=0.65\linewidth]{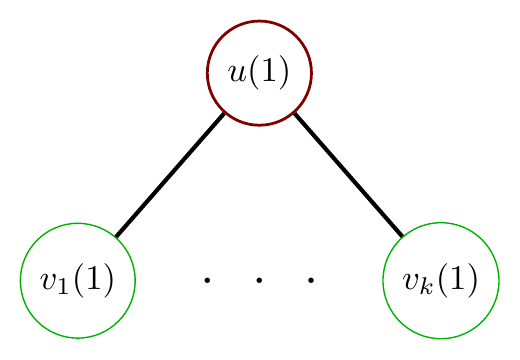}
            \caption{A star graph with $k$ leaves, featured over a single-value input-feature domain.}
            \label{figure:simple_star_sv}
    \end{center}
\end{minipage} \hfil\begin{minipage}{.32\linewidth}
    \begin{center}
            
            \includegraphics[width=0.65\linewidth]{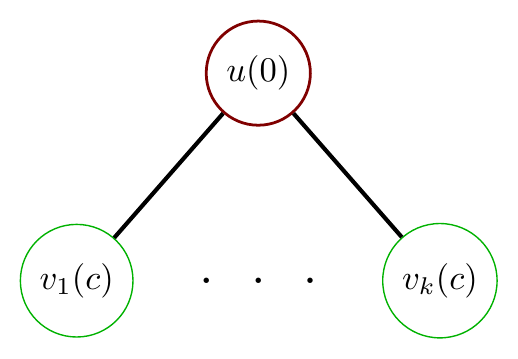}
            \caption{A star graph with $k$ leaves, featured over $\PNat$.}
            \label{figure:simple_star_uc}
    \end{center}
\end{minipage}\hfil\begin{minipage}{.32\linewidth}
        \begin{center}
            \includegraphics[width=0.65\linewidth]{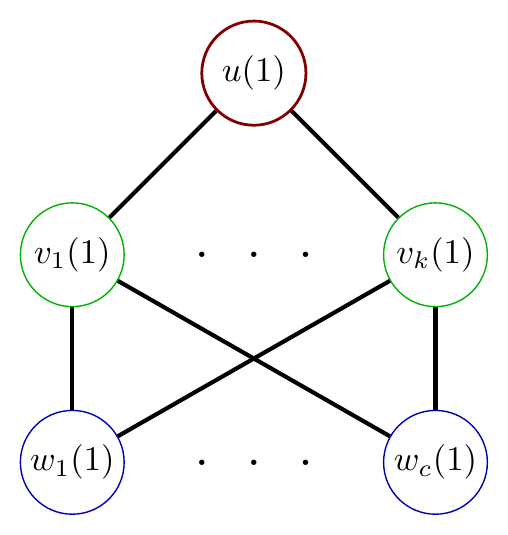}
            \caption{A tripartite graph, with $k$ intermediates fully connected to $c$ leaves, featured over a single-value input-feature domain.}
            \label{figure:two_level_sv}
        \end{center}
\end{minipage}
\end{figure*}

It has already been stated that \sags can express functions that \aags and \mxags cannot \cite{xu2018powerful}.
For the sake of completeness we provide formal proofs that \aags and \mxags subsume neither \sags nor each other.

\subsection{Mean and Max do not subsume Sum}\label{subsec:sum_not_subsumed}
Neither \aags nor \mxags subsume \sagsns, even when the input-feature domain is a single value.

We define a featured star graph with (a parameter) $k$ leaves, $G_k$ (see \Cref{figure:simple_star_sv}): For every $k\in\PNat$:
\begin{itemize}
    \item $V(G_k)=\{u\}\cup\{v_1,\ldots,v_k\}$
    \item $E(G_k)=\bigcup_{i\in [k]}\{\{u,v_i\}\}$
    \item $Z(G_k)=\{(u,1)\}\bigcup_{i\in [k]}\{(v_i,1)\}$
\end{itemize}
Let $\CN$ be an $m$-layer GNN. We define $u^{(t)}_k\coloneqq\CN^{(t)}(G_k,u)$, the feature of $u\in V(G_k)$ after operating the first $t$ layers of $\CN$. Note that $u^{(m)}_k=\CN(G_k,u)$.

\begin{lemma}\label{lemma:mean_max_cannot_sum}
Assume $\CN$ is a \aag or a \mxag. Let the maximum input dimension of any layer be $d$, and let the maximum Lipschitz-Constant of any FNN of $\CN$ be $a$. Then, for every $k$ it holds that $\abs{u^{(m)}_k}\leq (da)^m$.
\end{lemma}
\begin{theorem}\label{theorem:sum_strong}
Let $f:\CG_{1}\rightarrow\CZ_\Real$ a feature transformation such that for every $k$ it holds that $f(G_k)(u)=k$. Then, ${\text{\aags}\not\approx f}$ and \mxags $\not\approx f$.
\end{theorem}

Note that by \Cref{theorem:sum_strong}, a function such as neighbors-count is inexpressible by \aags and \mxags.

\begin{corollary}\label{corola:mean_max_no_sum}
We have that \aags $\nsubsumesp{\{1\}}$ \sagsns, \mxags $\nsubsumesp{\{1\}}$ \sagsns.
\end{corollary}

\subsection{Mean and Max do not subsume each other}
\aags and \mxags do not subsume each other, even in a finite input-feature domain setting. We define a parameterized graph in which, depending on the parameters' arguments, the average of the center's neighbors is in $[0,\frac{1}{2}]$ while their max can be either $0$ or $1$. For every $k\in\Nat$ and $b\in\{0,1\}$:
\begin{itemize}
    \item $V(G_{k,b})=\{u\}\cup\{v_1,\ldots,v_k\}\cup\{w\}$
    \item $E(G_{k,b})=\bigcup_{i\in [k]}\{\{u,v_i\}\}\cup\{\{u,w\}\}$
    \item $Z(G_{k,b})=\{(u,0)\}\bigcup_{i\in [k]}\{(v_i,0)\}\cup\{(w,b)\}$
\end{itemize}
\begin{theorem}\label{theorem:mean_strong}
Let $f:\CG_{\{0,1\}}\rightarrow\CZ_\Real$ a feature transformation such that for every $k$ it holds that $f(G_{k,b})(u)=\frac{b}{k+1}$. Then, \mxags$\not\approx f$.
\end{theorem}

\begin{theorem}\label{theorem:max_strong}
Let $f:\CG_{\{0,1\}}\rightarrow\CZ_\Real$ a feature transformation such that for every $k$ it holds that $f(G_{k,b})(u)=b$. Then, \aags$\not\approx f$.
\end{theorem}
		   
\begin{corollary}\label{corola:mean_max_no_other}
We have that \aags $\nsubsumesp{\{0,1\}}$ \mxags, \mxags $\nsubsumesp{\{0,1\}}$ \aags.
\end{corollary}

\section{Sometimes Sum Subsumes}\label{sec:sum_enough}
In a bounded input-feature domain setting, \sags can express every function that \aags and \mxags can. The bounded input-feature domain results in a bounded range for Mean and Max, a fact which can be exploited to approximate the target GNN with a Sum-GNN. The approximating Sum-GNNs, that we describe, come at a size cost. We do not know if an asymptotically-lower-cost construction exist.

\subsection{Mean by Sum}\label{subsec:mean_sum}
\sags subsume \aags in a bounded input-feature domain setting.

\begin{lemma}\label{lemma:mean_approximation_with_sum}
For every $\epsilon > 0$ and $d\in\PNat$, there exists a \sag $\CN$ of size $O(d\frac{1}{\epsilon})$ such that for every featured graph ${G\in\CG_{[0,1]\subset\Real^d}}$ it holds that ${\forall v\in V(G)}\; \abs{N(G,v)-\mean(v)}\leq \epsilon$.
\end{lemma}

\begin{theorem}\label{theorem:mean_sum_emulation}
Let a \aag $\CN_M$ consisting of $m$ layers, let the maximum input dimension of any layer be $d$, and let the maximum Lipschitz-Constant of any FNN of $\CN_M$ be $a$. Then, for every $\epsilon>0$ there exists a \sag $\CN_S$ such that:
\begin{itemize}
    \item [1.] ${\forall G\in\CG_{[0,1]^d}\; \forall v\in V(G) \quad |\CN_M(G,v)-\CN_S(G,v)|\leq\epsilon}$.
    \item [2.] $\abs{\CN_{S}}\leq O(\abs{\CN_M}+\frac{d\cdot m\cdot ad(1-(2ad)^m)}{\epsilon(1-(2ad))})$.
\end{itemize}
\end{theorem}
\begin{corollary}
\sags $\subsumesp{[0,1]}$ \aagsns.
\end{corollary}

\subsection{Max by Sum}\label{subsec:max_sum}
\sags subsume \mxags in a bounded input-feature domain setting.

\begin{lemma}\label{lemma:max_approximation_with_sum}
For every $\epsilon > 0$ and $d\in\PNat$, there exists a \sag $\CN$ of size $O(d\frac{1}{\epsilon})$ such that for every featured graph $G\in\CG_{[0,1]^d}$ and vertex $v\in V(G)$ it holds that $\abs{N(G,v)-\max(v)}\leq \epsilon$.
\end{lemma}

\begin{theorem}\label{theorem:max_sum_emulation}
Let a \mxag $\CN_M$ consisting of $m$ layers, let the maximum input dimension of any layer be $d$, and let the maximum Lipschitz-Constant of any FNN of $\CN_M$ be $a$. Then, for every $\epsilon>0$ there exists a \sag $\CN_S$ such that:
\begin{itemize}
    \item [1.] ${\forall G\in\CG_{[0,1]^d}\; \forall v\in V(G) \quad |\CN_M(G,v)-\CN_S(G,v)|\leq\epsilon}$.
    \item [2.] $\abs{\CN_{S}}\leq O(\abs{\CN_M}+\frac{d\cdot m\cdot ad(1-(2ad)^m)}{\epsilon(1-(2ad))})$.
\end{itemize}
\end{theorem}
\begin{corollary}
\sags $\subsumesp{[0,1]}$ \mxagsns.
\end{corollary}

\section{Mean and Max Have Their Place}\label{sec:mean_max_place}
In two important settings, Mean and Max aggregations enable expressing functions that cannot be expressed with Sum alone. As in \Cref{sec:sums_place}, we define a graph $G_\theta$ parameterized by $\theta$ over domain $\Theta$. We define a feature transformation $f$ on that graph and prove that it cannot be approximated by \sagsns. The line of proofs (in the appendix) is as follows:
\begin{itemize}    
    \item [1.] We show that for every \sag $\CN$ there exists a finite set $F_{\CN}$ of polynomials of $\theta$, those polynomials obtain a certain property $\phi$, and it holds that: 
    $${\forall \theta\in \Theta\ \exists u_{\theta}\in V(G_{\theta})\ \exists p\in F_{\CN}:\ \CN(G_{\theta},u_{\theta})=p(\theta)}$$
    \item [2.] We show that for every finite set $F$ of polynomials ${\text{(of }\theta\text{)}}$ that obtain $\phi$, it holds that:
    $${\forall \epsilon>0\ \exists \theta\in \Theta:\ \forall p\in F\ \abs{p(\theta)-f(G_{\theta})(u_{\theta})}>\epsilon}$$    
    
\end{itemize}

\subsection{Unbounded, Countable, Input-Feature Domain}\label{subsec:unbounded}
In an unbounded input-feature domain setting, Mean;Max and other GNNs are not subsumed by \sagsns. We define a graph $G_{k,c}$ (see \Cref{figure:simple_star_uc}): For $(k,c)\in\PNat^2$,
\begin{itemize}
    \item $V(G_{k,c})=\{u\}\cup\{v_1,\ldots,v_k\}$
    \item $E(G_{k,c})=\bigcup_{i\in [k]}\{\{u,v_i\}\}$
    \item $Z(G_{k,c})=\{(u,0)\}\bigcup_{i\in [k]}\{(v_i,c)\}$
\end{itemize}

\begin{theorem}\label{theorem:no_approx_unbounded}
Let $f:\CG_{\Nat^1}\rightarrow\CZ_\Real$ a feature transformation,
such that for every $k,c$ it holds that $f(G_{k,c})(u)=c$. Then, $\text{\sags }\not\approx f$.
\end{theorem}

\begin{corollary}\label{corola:not_dominant_nat}
Denote by $S$ the set of all multisets over $\PNat$. Let $g:S\rightarrow\Real$ an aggregation such that $\forall a,b\in\PNat\ g({\{a\} \choose b})=a$, that is, $g$ aggregates every homogeneous multiset to its single unique value.
Then, \sags $\nsubsumesp{\Nat}$ g-aggregation GNNs.
\end{corollary}

\Cref{corola:not_dominant_nat} implies a limitation of \sags compared to GNNs that use Mean; Max; or many other aggregations.

\subsubsection{Graph Embedding}
\sags are limited compared to Mean; Max; and other GNNs, not only when used to approximate vertices' feature transformations but also when used in combination with a readout function to approximate graph embeddings. Consider another variant of $G_{k,c}$: For $(k,c)\in\PNat^2$,
\begin{itemize}
    \item $V(G_{k,c})=\{u_1,\ldots,u_{k^2}\}\cup\{v_1,\ldots,v_k\}$
    \item $E(G_{k,c})=\bigcup_{i\in [k^2], j\in[k]}\{\{u_i,v_j\}\}$
    \item $Z(G_{k,c})=\bigcup_{i\in [k^2]}\{(u_i,0)\}\bigcup_{i\in [k]}\{(v_i,c)\}$
\end{itemize}

\begin{theorem}\label{theorem:gre_no_approx}
Let $f:\CG_{\Nat^1}\rightarrow\Real$ a graph embedding
such that $\forall k,c\ f(G_{k,c})=\frac{kc}{k+1}$.
Let an aggregation ${\Fa\in\{\lsum,\mean\}}$ and an FNN $\FF$, and define a readout $\Fro\coloneqq f_\FF\circ \Fa$.
Then, ${\Fro\circ\text{\sags }\not\approx f}$.
\end{theorem}

\begin{corollary}\label{corola:gre_not_dominant_nat}
Denote by $S$ the set of all multisets over $\PNat$. Let $g:S\rightarrow\Real$ an aggregation such that $\forall a,b\in\PNat\ g({\{a\} \choose b})=a$. Let an aggregation ${\Fa\in\{\lsum,\mean\}}$ and an FNN $\FF$, and define a readout $\Fro\coloneqq f_\FF\circ \Fa$. Then, ${\Fro\circ\text{\sags }\nsubsumesp{\Nat} \mean \circ\ g\text{-GNNs}}$.    
\end{corollary}

We have shown that \sags do not subsume Mean and Max (and many other) GNNs. The setting though, consisted of an input-feature domain $\PNat$, that is, countable unbounded.

\subsection{Finite Input-Feature Domain}\label{subsec:simple}
Mean and Max aggregations are essential also when the input-feature domain is just a single value i.e. when the input is featureless graphs. We define a new graph $G_{k,c}$ (see \Cref{figure:two_level_sv}): For every $(k,c)\in\PNat^2$,
\begin{itemize}
    \item $V(G_{k,c})=\{u\}\cup\{v_1,\ldots,v_k\}\cup\{w_1,\ldots,w_c\}$
    \item $E(G_{k,c})=\bigcup_{i\in [k]}\{\{u,v_i\}\}\bigcup_{i\in[k],j\in [c]}\{\{v_i,w_j\}\}$
    \item $Z(G_{k,c})= \{(u,1)\}\bigcup_{i\in [k]}\{(v_i,1)\}\bigcup_{i\in [c]}\{(w_i,1)\}$
\end{itemize}

\begin{theorem}\label{theorem:no_approx_bounded}
Let $f:\CG_{1}\rightarrow\CZ_\Real$ a feature transformation, such that for every $k,c$ it holds that $f(G_{k,c})(u)=c$. Then, \sags $\not\approx f$.
\end{theorem}

\begin{corollary}\label{corola:not_dominant_one}
Denote by $S$ the set of all multisets over $\PNat$, and let $g:S\rightarrow\Real$ an aggregation such that $\forall a,b\in\PNat\ g({\{a\} \choose b})=a$. Then, \sags $\nsubsumesp{\{1\}}$ (Sum, g)-GNNs.
\end{corollary}

\Cref{corola:not_dominant_one} implies a limitation of \sags compared to stereo aggergation GNNs that combine Sum with Mean; Max; or many other aggregations. The limitation exists even when the input-feature domain consists of only a single value.

\subsubsection{Graph Embedding}
Completing the no-subsumption picture, \sags are not subsuming, in a 2-values input-feature domain setting, also when used in combination with a readout function to approximate graph embeddings. We define $G_{k,c}$: For every $(k,c)\in\PNat^2$,
\begin{itemize}
    \item $V(G_{k,c})=\{u_1,\ldots,u_{k^2}\}\cup\{v_1,\ldots,v_{k^3}\}\cup\{w_1,\ldots,w_{kc}\}$
    \item $E(G_{k,c})=\bigcup_{j\in[k^2],i\in [k^3]}\{\{u_j,v_i\}\}\bigcup_{i\in[k^3],j\in [kc]}\{\{v_i,w_j\}\}$
    \item $Z(G_{k,c})= \bigcup_{i\in[k^2]}\{(u_i,0)\}\bigcup_{i\in [k^3]}\{(v_i,0)\}\bigcup_{i\in [kc]}\{(w_i,1)\}$
\end{itemize}

\begin{theorem}\label{theorem:gre_no_approx_sv}
Let $f:\CG_{\{0,1\}^1}\rightarrow\Real$ a graph embedding
such that $\forall k,c\ f(G_{k,c})=\frac{(k^2+kc)kc}{k^3+k^2+kc}$.
Let an aggregation ${\Fa\in\{\lsum,\mean\}}$ and an FNN $\FF$, and define a readout $\Fro\coloneqq f_\FF\circ \Fa$.
Then, ${\Fro\circ\text{\sags }\not\approx f}$.
\end{theorem}

\begin{corollary}\label{corola:gre_not_dominant_nat_sv}
Denote by $S$ the set of all multisets over $\PNat$. Let ${g:S\rightarrow\Real}$ an aggregation such that 
${\forall a,b\in\PNat\ g({\{a\} \choose b})=a}$. Let an aggregation ${\Fa\in\{\lsum,\mean\}}$ and an FNN $\FF$, and define a readout $\Fro\coloneqq f_\FF\circ \Fa$. Then, ${\Fro\circ\text{\sags }\nsubsumesp{\{0,1\}} \mean \circ\ \text{(Sum, g)-GNNs}}$.
\end{corollary}

\section{Sum and More are Not Enough}\label{sec:more_sum}
In previous sections we showed that \sags do not subsume \aagsns and \mxags, by proving that they cannot express specific functions. In this section, rather than comparing different GNNs classes we focus on one broad GNNs class and show that it is limited in its ability to express any one of a certain range of functions.

Denote by $S$ the set of all multisets over $\Real$, and let an aggregation ${\Fa:S\rightarrow\Real}$. We say that $\Fa$ is a \emph{uniform polynomial aggregation} (UPA) if and only if for every homogeneous multiset ${\{x\} \choose b}, x\in\Real, b\in\PNat$ it holds that $\Fa({\{x\} \choose b})$ is either a polynomial of $x$ or a polynomial of $(bx)$. Note that Sum; Mean; and Max are all UPAs. We say that a GNN $\CN=(\CL^{(1)},\ldots,\CL^{(m)})$ is an \mupa (Multiple UPA) if and only if the aggregation input to each of its layers is defined by a series of UPAs. That is, $\CL^{(i)}=(\FF^{(i)},(\Fa^{(i)}_1,\ldots,\Fa^{(i)}_{b_i}))$, for some $b_i$ UPAs.

We define a parameterized graph $G_k$ (see \Cref{figure:simple_star_sv}): For every $k\in\PNat$:
\begin{itemize}
    \item $V(G_k)=\{u\}\cup\{v_1,\ldots,v_k\}$
    \item $E(G_k)=\bigcup_{i\in [k]}\{\{u,v_i\}\}$
    \item $Z(G_k)=\{(u,1)\}\bigcup_{i\in [k]}\{(v_i,1)\}$
\end{itemize}

\begin{lemma}\label{lemma:piecewise_polynomial}
Let $\CA$ an $m$-layer \mupa architecture, let $l$ be the maximum depth of any FNN in $\CA$, and let $d$ be the maximum in-degree of any node in any FNN in $\CA$. Then, there exists $r\in\Nat$ such that: for every GNN $\CN$ that realizes $\CA$ it holds that $\CN(G_k,u)$ is piecewise-polynomial (of $k$) with at most $((d+1)^l)^m$ pieces, and each piece is of degree at most $r$.
\end{lemma}
\Cref{lemma:piecewise_polynomial} implies that the architecture bounds (from above) the number of polynomial pieces, and their degrees, that make the function computed by any particular realization of the architecture.
With \Cref{lemma:piecewise_polynomial} at our disposal, we consider any feature transformation that does not converge to a polynomial when applied to $u\in V(G_k)$ and viewed as a function of $k$. We show that such a function is inexpressible by \mupasns.

\begin{theorem}\label{theorem:no_converge_no_approx}
Let $f:\CG_{1}\rightarrow\CZ_\Real$ a feature transformation, and define $g(k)\coloneqq f(G_k)(u)$. 
Assume that $g$ does not converge to any polynomial, that is, there exists $\epsilon>0$ such that for every polynomial $p$, for every $K_0$, there exists $k\geq K_0$ such that $\abs{g(k)-p(k)}\geq \epsilon$. Then, \mupasns$\not\approx f$.
\end{theorem}

The last inexpressivity property we prove, concerns a class of functions which we call \emph{PIL} (Polynomial-Intersection Limited). For $n\in\Nat$ denote by $P_n$ the set of all polynomials of degree $\leq n$.
We say that a function $f:\mathbb{N}\rightarrow\mathbb{R}$ is PIL if and only if for every $n\in\Nat$ there exists $k_n\in\Nat$ such that for every polynomial $p\in P_n$ there exist at most $k_n-1$ consecutive integer points on which $p$ and $f$ assume the same value. Formally, $${\sup\big(k:\forall p\in P_n\; \forall x\in\Nat \; \forall y\in[x..(x+k-1)] \; f(y)=p(y)\big)\in\Nat}$$ We consider every feature transformation $f$ such that for $g(k)\coloneqq f(G_k)(u)$ it holds that $g$ is PIL. This is a different characterization than "no polynomial-convergence" (in \Cref{theorem:no_converge_no_approx}), and neither one implies the other. The result though, is weaker for the current characterization. We show that every \mupa architecture can approximate such a function only down to a certain $\epsilon>0$. 
That is, every GNN that realizes the architecture - no matter the specific weights of its FNNs - is far from the function by at least $\epsilon$ (at least in one point).
The following lemma is an adaptation of the Polynomial of Best Approximation theorem \cite{poly_of_best_approx2006,golomb1962lectures} to the integer domain. There, it is a step in the proof of the Equioscillation theorem attributed to Chebyshev \cite{wiki_equioscillation}.

\begin{lemma}\label{lemma:polynomial_approximation}
For $x,k\in\Nat$ define $I_{x,k}\coloneqq \{x,x+1,\ldots,x+k-1\}$ the set of consecutive $k$ integers starting at $x$.
Let $f:\mathbb{N}\rightarrow\mathbb{R}$ be a PIL, let $n\in\mathbb{N}$, and define $k_n\coloneqq$
$$1+\max(k:\forall p\in P_n\; \forall x\in\Nat \; \forall y\in[x..(x+k-1)] \; f(y)=p(y))$$
Then, for every $x\in\mathbb{N}$ there exists $\epsilon_{x,k_n}>0$ such that: for every $p\in P_n$ there exists $y\in I_{x,k_n}$ for which $\abs{p(y)-f(y)}\geq\epsilon_{x,k_n}$. That is, for every starting point $x$ there is a bounded interval $I_{x,k_n}$, and a gap $\epsilon_{x,k_n}$, such that no polynomial of degree $\leq n$ can approximate $f$ on that interval below that gap.
\end{lemma}

\begin{lemma}\label{lemma:polynomial_approximation_cont}
For every $q,n\in\Nat$ there exists a point $T_{q,n}\in\Nat$ and a gap $\delta_{T_{q,n}}>0$ such that: for every PIL $f:\mathbb{N}\rightarrow\mathbb{R}$, and every piecewise-polynomial $g$ with $q$ many pieces of degree $\leq n$, there exists $y\in\mathbb{N},\; 0\leq y\leq T_{q,n}$ for which $\abs{g(y)-f(y)}\geq\delta_{T_{q,n}}$. That is, the number of pieces and the max degree of a piecewise-polynomial $g$ determine a guaranteed minimum gap by which $g$ misses $f$ within a guaranteed interval.
\end{lemma}

\begin{theorem}\label{theorem:no_poly_limited_approx}
Let $f:\CG_{1}\rightarrow\CZ_\Real$ a feature transformation, let $g(k)\coloneqq f(G_k)(u)$, and assume that $g$ is PIL. Then, for every \mupa architecture $\CA$, there exists $\epsilon_{\CA}>0$ such that for every \mupa $\CN$ that realizes $\CA$ there exists $k$ such that $\abs{\CN(G_{k},u)-f(G_{k})(u)}\geq\epsilon$.
\end{theorem}

%% file: sum_expressivity_experiments.tex
\section{Experimentation}\label{sec:experiment}
\input{figures/exp_figure.tex}
We experiment with vertex-level regression tasks.
In previous sections we formally proved certain expressivity properties of Sum; Mean; and Max GNNs. Our goal in experimentation is to examine how these properties may affect practical learnability: searching for an approximating GNN using stochastic gradient-descend.
With training data ranging over only a small subsection of the true-distribution range, does the existence of a uniformly-expressing GNN increase the chance that a well-generalizing GNN will be learned?

Specific details concerning training and architecture, as well additional illustrations and extended analysis, can be found in the appendix \footnote{code for running the experiments is found at \url{https://github.com/toenshoff/Uniform_Graph_Learning}}.

\subsection{Data and Setup}
For the graphs in the experiments, and with our GNN architecture consisting of two GNN layers (see appendix), Mean and Max aggregations output the same value for every vertex, up to machine precision. Thus, it is enough to experiment with Mean and assume identical results for Max.

We conduct experiments with two different datasets, one corresponds to the approximation task in \Cref{subsec:unbounded}, and the other to the task in \Cref{subsec:simple}:
\begin{enumerate}
    \item \textbf{U}nbounded \textbf{C}ountable Feature Domain (UC): This dataset consists of the star graphs $\{G_{k,c}\}$ from \Cref{subsec:unbounded}, for $k,c\in[1..1000]$.
    The center's ground truth value is $c$, and it is the only vertex whose value we want to predict.
    
    \item \textbf{S}ingle-\textbf{V}alue Feature Domain (SV): This dataset consists of the graphs $\{G_{k,c}\}$ from \Cref{subsec:simple}, for $k,c\in[1..1000]$.
    Again, the center's ground truth value is $c$, and we do not consider the other vertices' predicted values.    
\end{enumerate}

As training data, we vary $k\in[1..100]$ and $c\in[1..100]$.
We therefore train on 10K graphs in each experiment.
Afterwards, we test each GNN model on larger graphs with $k\in[101..1000\}$ and $c\in[101..1000]$.
Here, we illustrate our results for two representing values of $k$: $500,1000$, for all values of $c$. Illustrations of the full results can be found in the appendix.
The increased range of $k$ and $c$ in testing simulates the scenario of unbounded graph sizes and unbounded feature values, allowing us to study the performance in terms of uniform expressivity with unbounded features. 

\subsection{Results}
Our primary evaluation metric is the relative error.
Formally, if $y_\text{pred}$ is the prediction of the GNN for the center vertex of an input graph $G$, with truth label $c$, we define the relative error as
\begin{equation*}
    \text{RE}\left(y_\text{pred}, c\right) = \frac{|y_\text{pred} - c|}{|c|}.
\end{equation*}
A relative error greater or equal to 1 is a strong evidence for inability to approximate, as the assessed approximation is no-better than an always-0 output.
It is also reasonable that in practice, when judging the regression of a function whose range vary by a factor of 1000, relative error would be the relevant measure.

\subsubsection{Unbounded, Countable, Feature Domain}
Figure \ref{fig:ucf} provides the test results for UC.
We plot the relative error against different values of $c$.
Note that the error has a logarithmic scale.
\aags achieve very low relative errors of less than $10^{-4}$ across all considered combinations of $k$ and $c$.
Their relative error falls to less than $10^{-6}$ when $c$ is within the range seen during training ($\leq100$),
Therefore, \aags do show some degree of overfitting. Notably, the value of $k$ has virtually no effect on the error of \aags.
This is expected, since mean aggregation should not be affected by the degree $k$ of a center vertex whose neighbors are identical, up to machine precision. \sags yield a substantially higher relative error.
For $k=500$ and $c\leq100$ the relative error is roughly $1$, but this value increases as $c$ grows beyond the training range. Crucially, the relative error of \sags also increases with $k$.
For $k=1000$, the relative error is above $1$ even when $c$ is within the range seen during training.
Therefore, \sags do generalize significantly worse than \aags in both parameters $k$ and $c$.
`
\subsubsection{Single-Value Feature Domain}
Figure \ref{fig:svf} provides the test results for SV.
Again, we plot the relative error against different values of $c$.
\sags yield similar relative errors as in the UC experiment.
As expected, learned (Sum,Mean)-GNNs do perform significantly better than \sagsns.
However, the learning of (Sum,Mean)-GNNs is not as successful as the learning of  \aags in the UC experiment: relative error is around $10^{-1}$ for $k=500$, and slightly larger for $k=1000$, clearly worse than the UC-experiment performance. In particular, the learned (Sum,Mean)-GNN is sensitive to increases in $k$. Note that each (Sum,Mean)-GNN layer receives both Sum and Mean aggregations arguments and needs to choose the right one, thus it is a different learning challenge than in the first experiment.

%% file: figures/exp_figure.tex
\begin{figure*}
\centering
\begin{subfigure}{.47\textwidth}
  \centering
  \includegraphics[width=.99\linewidth]{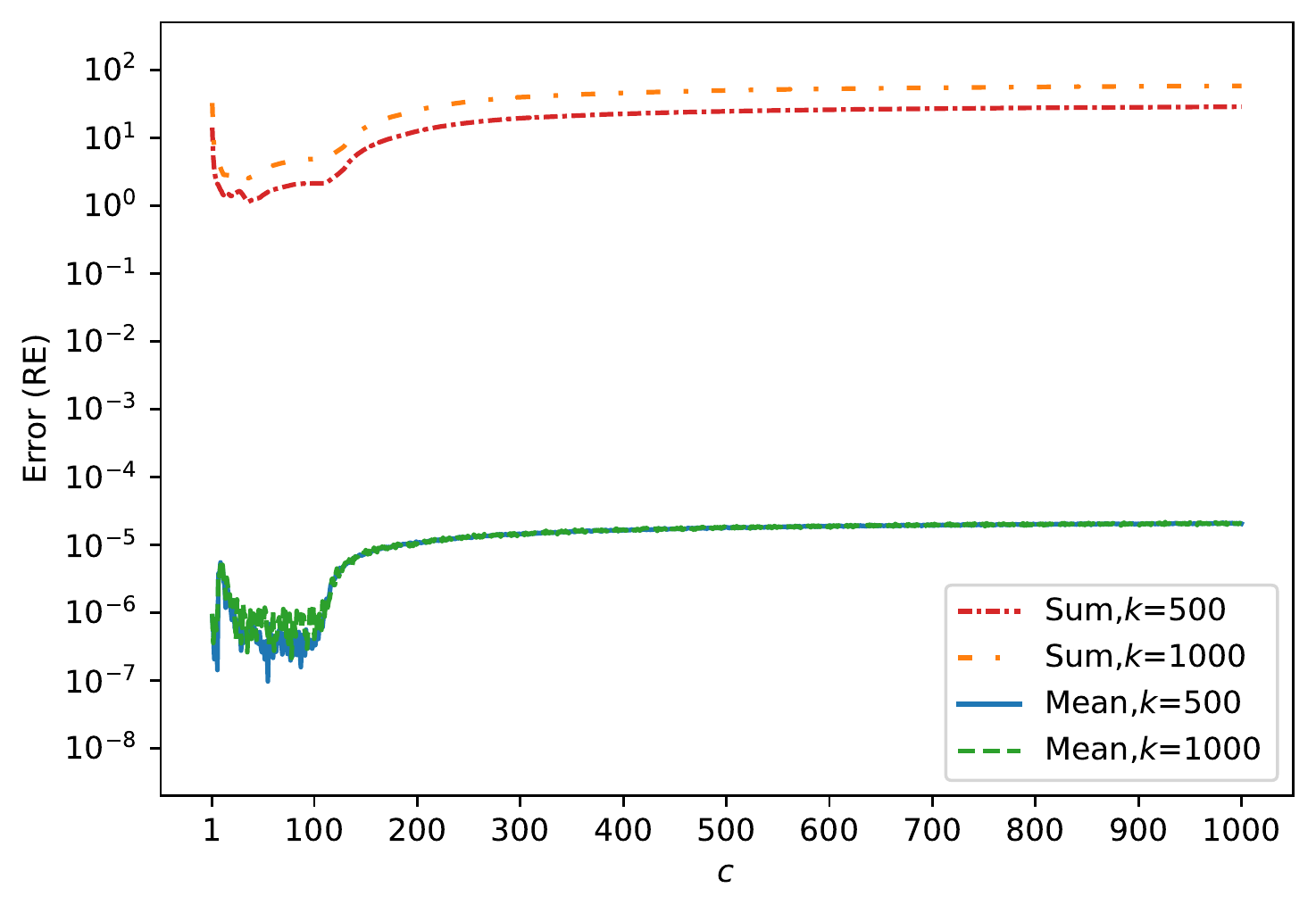}
  \caption{Unbounded Countable Features}
  \label{fig:ucf}
\end{subfigure}%
\hfill
\begin{subfigure}{.47\textwidth}
  \centering
  \includegraphics[width=.99\linewidth]{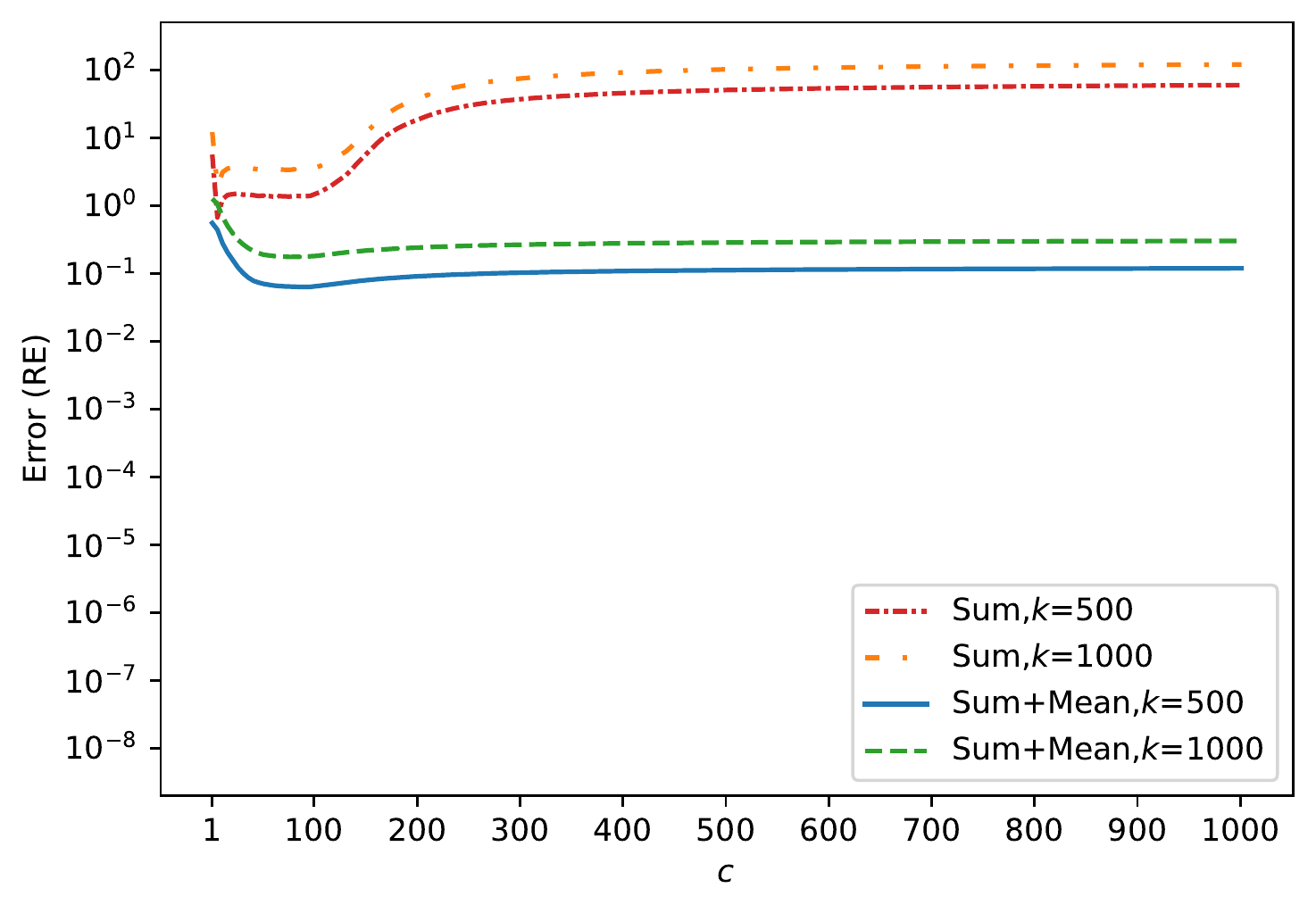}
  \caption{Single Value Features}
  \label{fig:svf}
\end{subfigure}
\caption{Relative Error of different aggregations on UC and SV.}
\label{fig:exp}
\end{figure*}

%% file: sum_expressivity_appendix.tex
\appendix
\section{Proofs}

For the reader's convenience, we re-state the results that are proven in this appendix.

\subsection*{Proofs for \Cref{sec:sums_place}}
\subsubsection{\Cref{lemma:mean_max_cannot_sum}}
																										
\textit{
Assume $\CN$ is a \aag or a \mxag. Let the maximum input dimension of any layer be $d$, and let the maximum Lipschitz-Constant of any FNN of $\CN$ be $a$. Then, for every $k$ it holds that $\abs{u^{(m)}_k}\leq (da)^m$.
}
\begin{proof}
For every $i,j\in[k]$ there is an 
automorphism of $G_k$ that maps $v_i$ to $v_j$, thus they receive the same feature throughout the computation. We define $v^{(t)}_k\coloneqq\CN^{(t)}(G_k,v_i)$ for every ${i\in[k]}$. We view $u^{(t)}_k, v^{(t)}_k$ as functions of $k$.
First, assume Assume $\CN$ is a \aag. We show by induction that for any $t\in [m]$ it holds that $\abs{v^{(t)}_k}\leq (2da)^t, \abs{u^{(t)}_k}\leq (2da)^t$. For $t=1$, $v^{(t)}_k=f_1(1,1)$ for some FNN $f_1$ whose Lipschitz-Constant is at most $a$, hence $\abs{f_1(1,\frac{1}{1})}\leq2a$. Also, $u^{(t)}_k=f_1(1,\frac{k}{k})=f_1(1,1)\leq2a$. Assume correctness for $t=n$. For $t=n+1$ we have $v^{(n+1)}_k=f_{n+1}(v^{(n)}_k,u^{(n)}_k)$ for some FNN $f_{n+1}$ whose Lipschitz-Constant is at most $a$. Hence, $v^{(n+1)}_k\leq 2da(2da)^{n}=(2da)^{n+1}$. 
Also, $u^{(n+1)}_k=f_{n+1}(u^{(n)}_k,\frac{kv^{(n)}_k)}{k})=f_{n+1}(u^{(n)}_k,v^{(n)}_k)\leq2da(2da)^{n}=(2da)^{n+1}$.

Next, assume $\CN$ is a \mxag. Notice that for every $t\in[0..(m-1)]$ it holds that 
$\frac{u^{(t)}_k}{1}=\max(u^{(t)}_k)$ and $\frac{kv^{(t)}_k}{k}=\max(v_k^{(t)},\ldots,v_k^{(t)})$. Hence, the proof idea for a \aag applies also for a \mxag.
\end{proof}

\subsubsection{\Cref{theorem:sum_strong}}
\textit{
Let $f:\CG_{1}\rightarrow\CZ_\Real$ a feature transformation such that for every $k$ it holds that $f(G_k)(u)=k$. Then, ${\text{\aags}\not\approx f}$ and \mxags $\not\approx f$.
}
\begin{proof}
Choose any $\epsilon>0$.
Let $\CN$ be either \aag or \mxag. Let the maximum input dimension of any layer be $d$, and let the maximum Lipschitz-Constant of any FNN of $\CN$ be $a$. Choose $k=(2da)^{m}+\epsilon$, then by \Cref{lemma:mean_max_cannot_sum} we have that $\abs{\CN(G_k,u)-f(G_k)(u)}\geq\epsilon$.
\end{proof}

\subsubsection{\Cref{corola:mean_max_no_sum}}
\textit{
We have that \aags $\nsubsumesp{\{1\}}$ \sagsns, \mxags $\nsubsumesp{\{1\}}$ \sagsns.
}
\begin{proof}
Clearly, there is a \sag that computes $f$ exactly. By \Cref{theorem:sum_strong}, there is no \aag or \mxag that approximates $f$.
\end{proof}

\subsubsection{\Cref{theorem:mean_strong}}
\textit{
Let $f:\CG_{\{0,1\}}\rightarrow\CZ_\Real$ a feature transformation such that for every $k$ it holds that $f(G_{k,b})(u)=\frac{b}{k+1}$. Then, \mxags$\not\approx f$.
}
\begin{proof}
Let $\CN$ be an $m$-layer \mxagsns. It is not hard to see by induction on $m$ that for every $i>0,j>0$ it holds that $\CN(G_{i,1},u)=\CN(G_{j,1},u)$. Hence, $\exists k:\abs{f(G_{k,1})(u)-\CN(G_{k,1},u)}>0.24$.
\end{proof}

\subsubsection{\Cref{theorem:max_strong}}
\textit{
Let $f:\CG_{\{0,1\}}\rightarrow\CZ_\Real$ a feature transformation such that for every $k$ it holds that $f(G_{k,b})(u)=b$. Then, \aags$\not\approx f$.
}
\begin{proof}
Let $\CN$ be an $m$-layer \aagsns. It is not hard to show that $N(G_{k,b},u)$ is Lipschitz-Continuous with respect to the aggregation and that with the aggregation being Mean we have that $\lim_{k\rightarrow\infty}\abs{N(G_{k,0},u)-N(G_{k,1},u)}=0$.
\end{proof}

\subsubsection{\Cref{corola:mean_max_no_other}}
\textit{
We have that \aags $\nsubsumesp{\{0,1\}}$ \mxags, \mxags $\nsubsumesp{\{0,1\}}$ \aags.
}
\begin{proof}
Clearly, there is a \aag that computes $f$ of \Cref{theorem:mean_strong} exactly, and by \Cref{theorem:mean_strong} there is no \mxag that approximates $f$. Clearly, there is a \mxag that computes $f$ of \Cref{theorem:max_strong} exactly, and by \Cref{theorem:max_strong} there is no \aag that approximates $f$.
\end{proof}

\subsection*{Proofs for \Cref{sec:sum_enough}}
Every reference in \Cref{lemma:avg_threshold_indicator} (and its proof) to a vertex-related value-vector is element-wise: for every vertex $v$ and a value-function $f(v)$ of output dimension $d$ we use the notation $f(v)$ to represent $f(v)_i$ for all $i\in[d]$.
\begin{lemma}\label{lemma:avg_threshold_indicator}
Let $d\in\PNat$, let $s\in[0,1]$, and let $0<a\leq s$. Then, there is a \sag $\CN$ such that for every featured graph $G\in\CG_{[0,1]^d}$ and every vertex $v\in V(G)$ it holds that
\[ 
\CN(G,v)= 
\begin{cases} 
      0 & s\leq \mean(v) \\
      
      \frac{n_v(s-\mean(v))}{a} &  s-\frac{a}{n_v} < \mean(v) < s\\
      
      1 & s-a\leq \mean(v) \leq s-\frac{a}{n_v} \\
      
      1-\frac{n_v(s-\mean(v)-a)}{a} &  s-a-\frac{a}{n_v} < 
      \\&\quad\mean(v) < s-a\\\\
      
      0 & \mean(v)\leq s-a-\frac{a}{n_v}
      
      \end{cases}
\]
\end{lemma}

\begin{figure*}[h]

\begin{minipage}[t]{.47\linewidth}
    \begin{center}
            \includegraphics[width=0.6\linewidth]{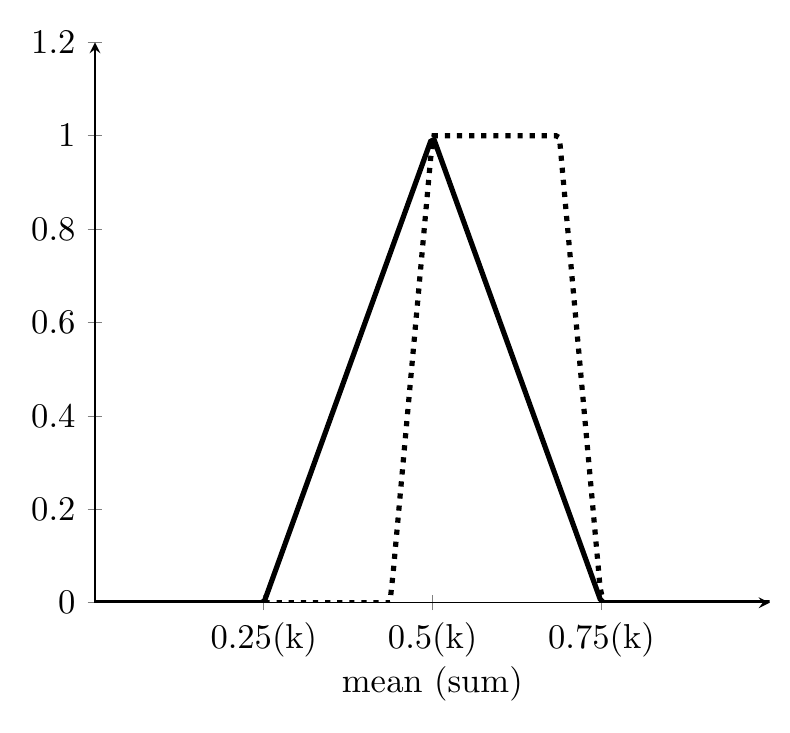}
            \caption{A single "position indicator", as constructed in \Cref{lemma:avg_threshold_indicator}, for interval $a=0.25$ and position $s=0.75$. The full line is for $n_v^-=1$ and the dotted line is for $n_v^-=4$. The x-axis represents $\mean(v)$ in the domain $[0,1]$ and the corresponding $\lsum(v)$ in the domain $[0,k]$.}
            \label{figure:figure_indicator}
    \end{center}
\end{minipage}\hfil\begin{minipage}[t]{.47\linewidth}
    \begin{center}
            \includegraphics[width=0.6\linewidth]{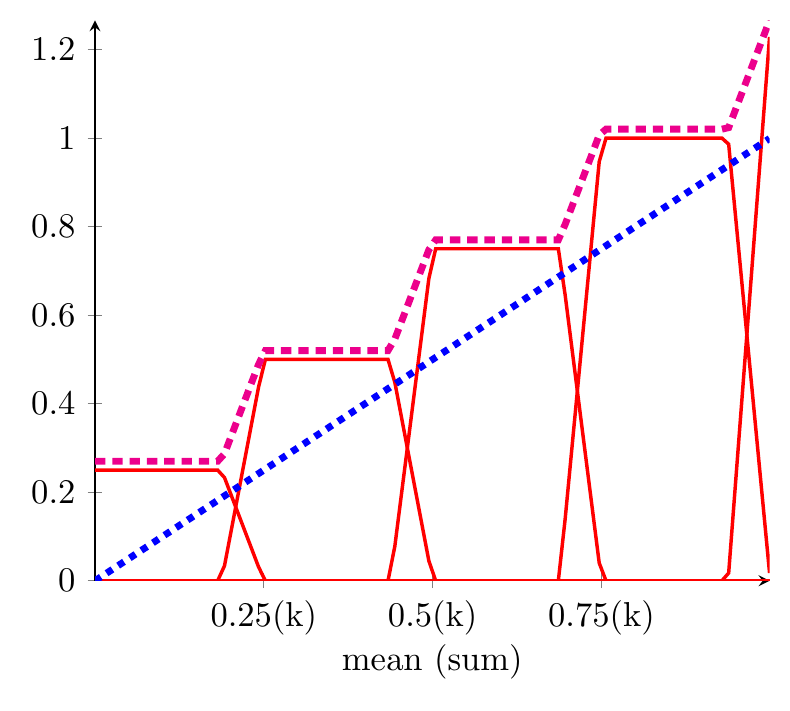}
            \caption{An illustration of the construction in \Cref{lemma:mean_approximation_with_sum}, for $a=0.25, n_v^-=4$. The solid red trapezoids are indicators scaled according to the position they are indicating. The dashed magenta steps-line is the sum of the indicators, which is the final function. The dotted blue line is the line to approximate. The x-axis represents $\mean(v)$ in the domain $[0,1]$ and the corresponding $\lsum(v)$ in the domain $[0,k]$.}
            \label{figure:mean_by_sum}
    \end{center}
\end{minipage}
\end{figure*}

\begin{proof}
Please refer to \Cref{figure:figure_indicator} for an illustration of the construction.
Let $v^{(t)}$ be the value of a vertex $v$ after layer $t$ and let $g^{(t)}_v=\sum_{w\in N(v)}w^{(t)}$ the sum of $v's$ neighbors' values after layer $t$. We denote the function computed in layer $t$ of $\CN$ by $f_t$, that is, $v^{(t)}=f_t(v^{(t-1)},g^{(t-1)}_v)$.
First, we map the value of a vertex (and the sum of its neighbors) to a 2-tuple with the first coordinate being $1$ and the second being the vertex' value. That is, we define $f_1:\mathbb{R}^2\rightarrow\mathbb{R}^2$ to be $f_1(x,y)=(1,x)$. 
Then, we define $f_2:\mathbb{R}^2\times\mathbb{R}^2\rightarrow\mathbb{R}$ to be $f_2(x,y)=\text{ReLU}(\frac{sy_1-y_2}{a})-\text{ReLU}(\frac{sy_1-y_2}{a}-1)+\text{ReLU}(\frac{sy_1-y_2}{a}-n_v-1)-\text{ReLU}(\frac{sy_1-y_2}{a}-n_v)$. That is, $v^{(2)}=\text{ReLU}(\frac{n_v(s-\mean(v))}{a})-\text{ReLU}(\frac{n_v(s-\mean(v))}{a}-1)+\text{ReLU}(\frac{n_v(s-\mean(v))}{a}-n_v-1)-\text{ReLU}(\frac{n_v(s-\mean(v))}{a}-n_v)$. To see why $v^{(2)}$ fulfills the requirements, we describe the values of each of the three components for the different ranges of $\mean(v)$.
\begin{itemize}

    \item $s\leq \mean(v)\Rightarrow \frac{n_v(s-\mean(v))}{a}\leq 0\Rightarrow
    \text{ReLU}(\frac{n_v(s-\mean(v))}{a})=\text{ReLU}(\frac{n_v(s-\mean(v))}{a}-1)=
    \text{ReLU}(\frac{n_v(s-\mean(v))}{a}-n_v)=\text{ReLU}(\frac{n_v(s-\mean(v))}{a}-n_v)=0
    \Rightarrow v^{(2)}=0$

    \item $s-\frac{a}{n_v} < \mean(v) < s\Rightarrow 0<\frac{n_v(s-\mean(v))}{a}< 1\Rightarrow \text{ReLU}(\frac{n_v(s-\mean(v))}{a})-\text{ReLU}(\frac{n_v(s-\mean(v))}{a}-1)=\frac{n_v(s-\mean(v))}{a};$ $\text{ReLU}(\frac{n_v(s-\mean(v))}{a}-n_v-1)=\text{ReLU}(\frac{n_v(s-\mean(v))}{a}-n_v)=0\Rightarrow v^{(2)}=\frac{n_v(s-\mean(v))}{a}$
    
    \item $s-a\leq \mean(v) \leq s-\frac{a}{n_v}\Rightarrow 1\leq\frac{n_v(s-\mean(v))}{a}\leq n_v\Rightarrow \text{ReLU}(\frac{n_v(s-\mean(v))}{a})-\text{ReLU}(\frac{n_v(s-\mean(v))}{a}-1)=1;$ $\text{ReLU}(\frac{n_v(s-\mean(v))}{a}-n_v-1)=\text{ReLU}(\frac{n_v(s-\mean(v))}{a}-n_v)=0
    \Rightarrow v^{(2)}_v=1$
    
    \item $s-a-\frac{a}{n_v} < \mean(v) < s-a\Rightarrow n_v<\frac{n_v(s-\mean(v))}{a}< n_v+1 \Rightarrow \text{ReLU}(\frac{n_v(s-\mean(v))}{a})-\text{ReLU}(\frac{n_v(s-\mean(v))}{a}-1)=1;$
    $\text{ReLU}(\frac{n_v(s-\mean(v))}{a}-n_v-1)=0;$
    $\text{ReLU}(\frac{n_v(s-\mean(v))}{a}-n_v)=\frac{n_v(s-\mean(v)-a)}{a}\Rightarrow h^{(2)}_v=1-\frac{n_v(s-\mean(v)-a)}{a}$
    
    \item $\mean(v) \leq s-a-\frac{a}{n_v}\Rightarrow n_v+1\leq\frac{n_v(s-\mean(v))}{a}\Rightarrow \text{ReLU}(\frac{n_v(s-\mean(v))}{a})-\text{ReLU}(\frac{n_v(s-\mean(v))}{a}-1)=1;
    \text{ReLU}(\frac{n_v(s-\mean(v))}{a}-n_v-1)+\text{ReLU}(\frac{n_v(s-\mean(v))}{a}-n_v)=1
    \Rightarrow h^{(2)}_v=0$
    
\end{itemize}
\end{proof}

\subsubsection{\Cref{lemma:mean_approximation_with_sum}}
\textit{
For every $\epsilon > 0$ and $d\in\PNat$, there exists a \sag $\CN$ of size $O(d\frac{1}{\epsilon})$ such that for every featured graph ${G\in\CG_{[0,1]\subset\Real^d}}$ it holds that ${\forall v\in V(G)}\; \abs{N(G,v)-\mean(v)}\leq \epsilon$.
}
\begin{proof}
Please refer to \Cref{figure:mean_by_sum} for an illustration of the construction.
We describe a construction of size $O(\frac{1}{\epsilon})$ which approximates Mean for one coordinate, the extension to $d$ is by a simple duplication.
Every reference to a vertex-related value-vector is element-wise: for every vertex $v$ and a value-function $f(v)$ of output dimension $d$, we use the notation $f(v)$ to represent $f(v)_i$ for all $i\in[d]$.

Let $q\in\mathbb{N}_{>0}$ be the minimal natural such that $\frac{1}{q}<\epsilon$, and define $a=\frac{1}{q}$. 
Define $\{s_1=a,s_2=2a,\ldots,s_{q+1}=1+a)\}$. The first layer of $\CN$ is identical to $f_1$ in the \Cref{lemma:avg_threshold_indicator}. The second layer uses a copy of $f_2$ from the \Cref{lemma:avg_threshold_indicator}, for each $s_i$, multiplied by $s_i$, and then sums the $q+1$ outputs. To see why this is correct, assume $s_i-a\leq \mean(v)\leq s_i$. For $j<i\text{\; or\; }j>i+1$ we have by \Cref{lemma:avg_threshold_indicator} zero contribution of $s_j$ to the final sum. Next, if $s_i-\frac{a}{n_v} \leq \mean(v)$ then by \Cref{lemma:avg_threshold_indicator} we have a contribution of $$s_{i+1}\Big(1-\frac{n_v(s_{i+1}-\mean(v)-a)}{a}\Big)+s_i\Big(\frac{n_v(s_i-\mean(v))}{a}\Big)=$$ $$s_{i+1}\Big(1-\frac{n_v(s_{i}-\mean(v))}{a}\Big)+s_i\Big(\frac{n_v(s_i-\mean(v))}{a}\Big)$$ Denoting the last term by $x$ and considering that $s_i-\frac{a}{n_v} \leq \mean(v)\leq s_i$ we have that $\mean(v)\leq x\leq \mean(v)+a$. Finally, if $\mean(v) \leq s-\frac{a}{n_v}$ then by \Cref{lemma:avg_threshold_indicator} we have zero contribution of $s_{i+1}$ and a contribution of $s_i\leq \mean(v)+a$. Overall, we have that $\mean(v)\leq \CN(G,v)\leq \mean(v)+a$.
\end{proof}

\subsubsection{\Cref{theorem:mean_sum_emulation}}
\textit{
Let a \aag $\CN_M$ consisting of $m$ layers, let the maximum input dimension of any layer be $d$, and let the maximum Lipschitz-Constant of any FNN of $\CN_M$ be $a$. Then, for every $\epsilon>0$ there exists a \sag $\CN_S$ such that:
\begin{itemize}
    \item [1.] ${\forall G\in\CG_{[0,1]^d}\; \forall v\in V(G) \quad |\CN_M(G,v)-\CN_S(G,v)|\leq\epsilon}$.
    \item [2.] $\abs{\CN_{S}}\leq O(\abs{\CN_M}+\frac{d\cdot m\cdot ad(1-(2ad)^m)}{\epsilon(1-(2ad))})$.
\end{itemize}
}
\begin{proof}
Let $\CN_M=((f_1,\text{Mean}),\ldots,(f_m,\text{Mean}))$, that is, $f_1,\ldots,f_m$ are the FNNs constituting $\CN_M$'s layers.
Let $\hat{\epsilon}>0$ and let $\CN_{\hat{\epsilon}}=((g_1,\text{Sum}),(g_2,\text{Sum}))$ the GNN constructed in \Cref{lemma:mean_approximation_with_sum}, with parameter $\hat{\epsilon}$. Note that $g_1$ is indifferent to the aggregation parameter and $g_2$ is indifferent to the vertex's state parameter, thus, for both parameters an argument of '0' is as good as any other. Define a \sag with $2m$ layers $\CN_S=((\hat{f_1},\text{Sum}),\ldots,(\hat{f}_{2m},\text{Sum}))$. For $j=0\ldots(m-1)$, each pair of layers $(\hat{f}_{2j+1},\text{Sum}),(\hat{f}_{2(j+1)},\text{Sum})$ approximates the operation of $(f_{j+1},\text{Mean})$.
For a graph $G$ and a vertex $v\in V(G)$, denote the feature of $v$ after the $(2(j+1))^{th}$ layer of $\CN_S$ by $\hat{v}^{(2(j+1))}$, with $\hat{v}^{(0)}\coloneqq Z(G)(v)$. We define $(\hat{f}_{2j+1},\hat{f}_{2(j+1)})$ as follows.
$$\hat{f}_{2j+1}(\hat{v}^{(2j)},\Sigma_{w\in N(v)}\hat{w}^{(2j)})\coloneqq (\hat{v}^{(2j)},g_1(\hat{v}^{(2j)},0))$$
$$\hat{f}_{2(j+1)}((\hat{v}^{(2j)},g_1(\hat{v}^{(2j)},0)),\Sigma_{w\in N(v)}(\hat{w}^{(2j)},g_1(\hat{w}^{(2j)},0))\coloneqq$$
$$f_{j+1}(\hat{v}^{(2j)},g_2(0,\Sigma_{w\in N(v)}g_1(\hat{w}^{(2j)},0)))$$
For $t\in[m]$ denote the feature of $v$ after the $t^{th}$ layer of $\CN_M$ by $v^{(t)}$, with $v^{(0)}\coloneqq Z(G)(v)$, and denote by $e_t\coloneqq\abs{\hat{v}^{(2t)}_i-v^{(t)}_i}$ the maximum error of any coordinate of the output of the $(2t)^{th}$ layer of $\CN_S$. We prove by induction on $t$ that $e_t\leq ad\hat{\epsilon}\Sigma_{i\in[t]}(2ad)^{i-1}$. Denote that upper bound by $b_t$.
For $t=1$, we have
$$e_{1}=\abs{\hat{v}^{(2)}-v^{(1)}}=
f_1(\hat{v}^{(0)},g_2(0,\Sigma_{w\in N(v)}g_1(\hat{w}^{(0)},0)))-$$
$$f_1(v^{(0)},\text{Mean}(\{w^{(0)} | w\in N(v) \})|$$
The first $d$ input coordinates to $f_1$ are identical. For each coordinate $i$ of the last $d$ coordinates, 
by definition of $g_1$ and $g_2$ we have
$$\abs{g_2(0,\Sigma_{w\in N(v)}g_1(\hat{w}^{(0)},0))_i-\text{Mean}(\{w^{(0)}:w\in N(v)\})_i}\leq \hat{\epsilon}$$That difference translates to a difference of at most $a\hat{\epsilon}$ in any coordinate of $\abs{\hat{v}^{(2)}-v^{(1)}}$. In total, we have $e_1\leq ad\hat{\epsilon}$. Assume correctness for $t=n$. Layer $2(n+1)$ of $\CN_S$ is, by definition, the operation of $f_{n+1}$ on at most $2\cdot d$ coordinates. The first $d$ coordinates constitute $\hat{v}^{(2n)}$ and the last $d$ coordinates constitute $g_2(0,\Sigma_{w\in N(v)}g_1(\hat{w}^{(2n)},0))$. The error of each of the first $d$ coordinates is, by assumption, at most $b_n$. For each coordinate $i$ of the last $d$ coordinates, we have by assumption $$\forall w\in N(v)\;\; \abs{\hat{w}^{(2n)}_i-w^{(n)}_i}\leq b_n$$ 
hence 
\begin{align}
\abs{\frac{1}{|N(v)|}\Sigma_{w\in N(v)}\hat{w}^{(2n)}_i-\frac{1}{|N(v)|}\Sigma_{w\in N(v)}w^{(n)}_i}\leq b_n
\end{align}

hence, by definition of $g_1$ and $g_2$,
\begin{align}
    \abs{g_2(0,\Sigma_{w\in N(v)}g_1(\hat{w}^{(2n)},0))_i - \frac{1}{|N(v)|}\Sigma_{w\in N(v)}w^{(n)}_i}\leq
    \\ \nonumber b_n + \hat{\epsilon}
\end{align}

Combining the error bounds for the two types of input, we have that
$$e_{n+1}=\max(\abs{\hat{v}^{(2(n+1))}_i-v^{(n+1)}_i} : i\in[d])=$$
$$\max(|f_{n+1}(\hat{v}^{(2n)},g_2(0,\Sigma_{w\in N(v)}g_1(\hat{w}^{(2n)},0)))_i-$$
$$f_{n+1}(v^{(n)},\text{Mean}(\{w^{(n)}:w\in N(v)\}))_i| : i\in[d])\leq$$
$$adb_n+ad(b_n+\hat{\epsilon})=2adb_n+ad\hat{\epsilon}=$$
$$ad\hat{\epsilon}\Sigma_{i=2}^{n+1}(2ad)^{i-1} + ad\hat{\epsilon}=$$
$$ad\hat{\epsilon}\Sigma_{i\in[n+1]}(2ad)^{i-1}$$
With the induction proven, we have that $$b_m=ad\hat{\epsilon}\Sigma_{i\in[m]}(2ad)^{i-1}=\hat{\epsilon}ad\frac{(1-(2ad)^m)}{1-2ad}$$
Hence, the requirement that $b_m\leq\epsilon$ can be satisfied by setting
$$\hat{\epsilon}=\epsilon \frac{1-2ad}{ad(1-(2ad)^m)}$$
implying $$\frac{1}{\hat{\epsilon}}=\frac{ad(1-(2ad)^m)}{\epsilon(1-2ad)}$$
Finally, using \Cref{lemma:mean_approximation_with_sum} we have that for each $i\in[m]$ it holds
$$\abs{\hat{f}_{2i-1}}+\abs{\hat{f}_{2i}}=O\Big(\frac{d\cdot ad(1-(2ad)^m)}{\epsilon(1-2ad)}\Big) + \abs{f_i}$$
hence
$$|\CN_S|=|\CN_M|+O\Big(\frac{m\cdot d\cdot ad(1-(2ad)^m)}{\epsilon(1-2ad)}\Big)$$
\end{proof}

\begin{lemma}\label{lemma:max_by_sum}
Let $q\in\PNat$, define $a\coloneqq\frac{1}{q}$, and define a function $f:[0,1]\rightarrow\mathbb{R}^q$ such that $$f(x)_i\coloneqq\max(0, min(x-a(i-1),a))$$ That is, $f$ is an almost-unary representation of $x$ in units of $\frac{1}{q}$, "almsot" because it may contain a fraction (between $0$  and $1$) in its last coordinate. For a finite multiset $x=\{x_1,\ldots,x_n\},x_i\in[0,1]$, define $$g(x)\coloneqq min((a,\ldots,a),\Sigma_{i\in[n]}f(x_i))$$ a mapping from the multiset to the sum of its elements' representation, coordinate-wise capped at $a$. Then, $$\max(x)\leq \Sigma_{i\in [q]}g(x)_i\leq \max(x)+a$$
\end{lemma}
\begin{proof}
w.l.o.g assume $\max(x)=x_1$. For the lower bound, it is not hard to verify that $\forall i\in[q]\; g(x)_i\geq f(x_1)_i$, hence $\Sigma_{i\in [q]}g(x)_i\geq\Sigma_{i\in [q]}f(x_1)_i=x_1$. For the upper bound, assume $j=\max(i:g(x)_i>0)$, then necessarily $x_1\geq(j-1)a$ and $\Sigma_{i\in [q]}g(x)_i\leq ja$, hence $\Sigma_{i\in [q]}g(x)_i\leq x_1+a$.
\end{proof}

\subsubsection{\Cref{lemma:max_approximation_with_sum}}
\textit{
For every $\epsilon > 0$ and $d\in\PNat$, there exists a \sag $\CN$ of size $O(d\frac{1}{\epsilon})$ such that for every featured graph $G\in\CG_{[0,1]^d}$ and vertex $v\in V(G)$ it holds that $\abs{N(G,v)-\max(v)}\leq \epsilon$.
}
\begin{proof}
We describe a construction of size $O(\frac{1}{\epsilon})$ that approximates Max for one coordinate, the extension to $d$ is by a simple duplication. Every reference to a vertex-related value-vector is element-wise: for every vertex $v$ and a value-function $f(v)$ of output dimension $d$, we use the notation $f(v)$ to represent $f(v)_i$ for all $i\in[d]$.

Let $q\in\mathbb{N}_{>0}$ be the minimal natural such that $\frac{1}{q}<\epsilon$ and define $a\coloneqq\frac{1}{q}$. The first GNN layer computes for each vertex $v$ a vector $v^{(1)}\in [0,a]^{q}$ such that $(v^{(1)})_i=ReLU(Z(v)-(i-1)a)-ReLU(Z(v)-(i-1)a-a)$. Observe that the computation corresponds to the mapping $f$ in \Cref{lemma:max_by_sum}. The second GNN layer first caps the sum-aggregation of the neighbors' vectors, then sums the coordinates of the capped vector. That is, for a vertex $v$, let $y_v=\Sigma_{w\in N(v)}w^{(1)}$, then $v^{(2)}=\Sigma_{i\in[q]}(ReLU((y_v)_i)-ReLU((y_v)_i-a))$. Using \Cref{lemma:max_by_sum}, we get that $\max(v)\leq v^{(2)}\leq \max(v)+a<\max(v)+\epsilon$.
\end{proof}

\subsubsection{\Cref{theorem:max_sum_emulation}}
\textit{
Let a \mxag $\CN_M$ consisting of $m$ layers, let the maximum input dimension of any layer be $d$, and let the maximum Lipschitz-Constant of any FNN of $\CN_M$ be $a$. Then, for every $\epsilon>0$ there exists a \sag $\CN_S$ such that:
\begin{itemize}
    \item [1.] ${\forall G\in\CG_{[0,1]^d}\; \forall v\in V(G) \quad |\CN_M(G,v)-\CN_S(G,v)|\leq\epsilon}$.
    \item [2.] $\abs{\CN_{S}}\leq O(\abs{\CN_M}+\frac{d\cdot m\cdot ad(1-(2ad)^m)}{\epsilon(1-(2ad))})$.
\end{itemize}
}
\begin{proof}
The proof is identical to the \Cref{theorem:mean_sum_emulation} with the following adaptations:
\begin{itemize}
    \item [1.] Replacing any mention of 'Mean', with 'Max'.
    \item [2.] Replacing any usage of \Cref{lemma:mean_approximation_with_sum}, with \Cref{lemma:max_approximation_with_sum}.
    \item [3.] Replacing equations (1),(2), with equations (3),(4) hereinafter.
\end{itemize}

\begin{equation}
    \begin{split}
    \Big |\max(\hat{w}^{(2n)}_i : w\in N(v)) &-
    \\\max(w^{(n)}_i : w\in N(v))\Big |&\leq b_n
    \end{split}
\end{equation}
\begin{align}
    \Big| g_2(0,\Sigma_{w\in N(v)}g_1(\hat{w}^{(2n)},0))_i -\\
    \max(w^{(n)}_i : w\in N(v))\Big|\leq b_n + \hat{\epsilon}\nonumber
\end{align}
\end{proof}

\subsection*{Proofs for \Cref{sec:mean_max_place}}
\subsection{Describability}\label{subsec:describability}
Let $F$ be a set of polynomials in $k,c$, and let $g(k,c)$ be a function in $k,c$.

We say that $F$ \emph{weakly-describes} $g$ if and only if:
\begin{itemize}
    \item [a.]$F$ is finite.
    \item [b.] $\forall k,c\in\Nat\ \ \exists p\in F\ :\ p(k,c)=g(k,c)$.  
\end{itemize}

We identify a polynomial $p(k,c)$ as being \emph{good} if and only if $p(k,c)=\Sigma_{i\in[n],j\in[n]}a_{i,j}k^ic^j +\Sigma_{i=0}^n b_ik^i$ for some real coefficients $\{a_{i,j}\},\{b_i\}$ and some maximum degree $n\in\Nat$. That is, $p(k,c)$ is a polynomial in $k,c$ with max degrees $n$ for $k,c$, and every appearance of $c$ is with multiplication by a polynomial of $k$ of degree at least $1$. We say that $F$ is \emph{good} if and only if every polynomial in it is good.

We say that $F$ \emph{describes} $g$ if and only if: $F$ weakly-describes $g$ and $F$ is good.
We say that $g$ is \emph{describable} (\emph{w-describable}) if and only if there exists a set that (weakly-) describes it.

Let $F$ be a finite set of polynomials in $k,c$, we denote by ${\CB(F)\coloneqq \{k^ic^j\ :\ \exists p\in F\quad p=(...+a_{i,j}k^ic^j)\quad a_{i,j}\neq 0\}}$ the building blocks of $F$, that is, the degree combinations that appear in any of the polynomials in $F$. 
Let $b\in\{k,c\}$, we define ${b\CB(F)\coloneqq\{b\cdot k^ic^j : k^ic^j\in\CB(F)\}}$.

For every $a\in\Real$ and a set of functions $F$ of $k,c$, we define $aF\coloneqq\{af:f\in F\}$, and $(a+F)\coloneqq\{a+f:f\in F\}$. For two sets of functions $F,H$ of $k,c$, we define $F+H\coloneqq \{f+h : f\in F, h\in H\}$.
\begin{lemma}\label{lemma:describe_preserve}
\begin{itemize}
    \item [a.] Let $f(k,c)$ a function (w-)describable by a set $F$. Let $g(k,c):=ReLU(f(k,c))$ the composition of $ReLU$ over $f(k,c)$, then $g$ is (w-)describable by a set $F'$ such that $\CB(F')\subseteq (\CB(F)\cup\{k^0c^0\})$.
    \item [b.] Let $f_1(k,c),\ldots,f_l(k,c)$ be functions (w-)describable by $F_1,\ldots,F_l$ respectively. Then, for every real coefficients $\{a_i\},b$ the affine function $(\Sigma_{i=1}^n a_if_i)+b$ is (w-)describable by a set $F$ such that $\CB(F)\subseteq (\{k^0c^0\}\bigcup_{i\in[l]}\CB(F_i))$.
    \item [c.] Each output of a ReLU activated FNN whose inputs are all (w-)describable by a set $F$ is (w-)describable by a set $F'$ such that $\CB(F')\subseteq (\CB(F)\cup\{k^0c^0\})$.
    \item [d.] Let $f(k,c)$ a function w-describable by a set $F$, then $kf(k,c)$ is describable by some set $F'$ such that $\CB(F')\subseteq k\CB(F)$, and $cf(k,c)$ is w-describable by a set $F''$ such that $\CB(F'')\subseteq c\CB(F)$.
\end{itemize}
\end{lemma}
\begin{proof}
a. Let $F$ a set that (w-)describes the function $f$. For any $k,c$ either $g(k,c)=f(k,c)$ or $g(k,c)=0$, hence $ReLU(f)$ is (w-)describable by $F\cup\{0\}$.
\\b. It is not hard to verify that if $f_i$ is (w-)describable by $F_i$ then for every $a\in\mathbb{R}$ it holds that $af_i$ is (w-)describable by $aF_i$, and $f_i+a$ is (w-)describable by $F_i+a$. It is also not hard to verify then that for any $a_i,a_j\in\mathbb{R}$ it holds that $a_if_i+a_jf_j$ is (w-)describable by $(a_iF_i)+(a_jF_j)$. A straightforward induction proves that a linear combination of arbitrarily many (w-)describable functions is (w-)describable. Finally, let $F$ a set that (w-)describes the linear combination, then $F+b$ is a set that (w-)describes the affine function.
\\c. Implied by (a)+(b).
\\d. It is not hard to verify that if $f$ is w-describable by $F$ then $kf$ is describable by $kF$. Also, it is not hard to verify that if $f$ is w-describable by $F$ then $cf$ is w-describable by $cF$.
\end{proof}

\begin{lemma}\label{lemma:one_of_finite}
Let a series of graphs $\{H_{k,c}\}$, parametarized in $k,c\in\PNat$, each having an identified vertex $u$, such that for every $m$-layer \sag $\CN$ it holds that $\CN(H_{k,c},u)$, viewed as a function of $k,c$, is describable. Then, for every \sag $\CN$ and for every $\epsilon>0$ there exist $k,c$ s.t $\abs{\CN(H_{k,c},u)-c}>\epsilon$.
\end{lemma}
\begin{proof}
Let $F$ be a finite set of polynomials that describes $\CN(H_{k,c},u)$.
Fix any specific $c\in\PNat$, and for $K\in\PNat$ denote by $F_{K,c}=\{p\in F: \exists k\geq K: \CN(H_{k,c},u)=p(k,c)\}$
only those polynomials in $F$ that intersect with  $u^{(m)}_{k,c}$ in the domain $[K,\infty)\times \{c\}$. 
Denote the polynomials in $F_{K,c}$ that are a constant, by $\widehat{F}_{K,c}=\{p:p\in F_{K,c},\ p \text{ constant}\}$.
Let $\epsilon>0$ and assume by contradiction that for every $k\in\PNat$ it holds that $\abs{\CN(H_{k,c},u)-c}\leq\epsilon$. Then, there must exist $K_c\in\PNat$ for which $\widehat{F}_{K_c,c}=F_{K_c,c}$. Otherwise, as $F$ is assumed to describe $\CN(H_{k,c},u)$, any appearance of $c$, in any $p\in F_{k,c}$, is tied to $k$, and we would have $$\inf_{p\in (F_{k,c}\setminus \widehat{F}_{k,c})} \abs{p(k,c)}\xrightarrow[k\rightarrow\infty]{}\infty$$ and $$\sup_{k\in \Nat} \abs{\CN(H_{k,c},u)-c}=\infty$$ in contradiction to $\abs{\CN(H_{k,c},u)-c}\leq\epsilon$.
By definition, $\widehat{F}_{K,c}$ is a subset of $F$ which is finite, and so $\max(\widehat{F}_{K_c,c})\leq \max(p\in F : p \text{ constant})$.
Denote the last term by $max_F$. As our reasoning thus far is true for any $c$, it holds that $\max(\max(\widehat{F}_{K_c,c}):{c\in\Nat})\leq max_F$. Finally, for $c=\ceil{max_F+\epsilon+1}$ necessarily for all $k\geq K_c$ it holds that 
$\abs{\CN(H_{k,c},u)-c}>c-max_F>\epsilon$.
\end{proof}

\subsection*{\Cref{subsec:unbounded}}
Define a series of featured star graphs $\{G_{k,c}\}$ as follows: For $(k,c)\in\PNat^2$,
\begin{itemize}
    \item $V(G_{k,c})=\{u\}\cup\{v_1,\ldots,v_k\}$
    \item $E(G_{k,c})=\bigcup_{i\in [k]}\{\{u,v_i\}\}$
    \item $Z(G_{k,c})=\{(u,0)\}\bigcup_{i\in [k]}\{(v_i,c)\}$
\end{itemize}

Let $\CN$ be an $m$-layer \sagns. We define $u^{(t)}_{k,c}\coloneqq\CN^{(t)}(G_{k,c},u)$, the feature of $u\in V(G_{k,c})$ after operating the first $t$ layers of $\CN$. Note that $u^{(m)}_{k,c}=\CN(G_{k,c},u)$. For every $i,j\in[k]$ there is an automorphism of $G_k$ that maps $v_i$ to $v_j$, thus they receive the same feature throughout the computation. We define $v^{(t)}_{k,c}\coloneqq\CN^{(t)}(G_{k,c},v_i)$ for every ${i\in[k]}$. In our argumentation, we view $u^{(t)}_{k,c}, v^{(t)}_{k,c}$ as functions of $k,c$.
\begin{lemma}\label{lemma:describable}
It holds that $u^{(m)}_{k,c}$ is describable.
\end{lemma}

\begin{proof}
We show by induction that for every $t\in[m]$ it holds that $v_{k,c}^{(t)}$ is w-describable and that $u_{k,c}^{(t)}$ is describable. For $t=0$ we have $u_{k,c}^{(t)}=0, v_{k,c}^{(t)}=c$ and the assumption holds. Assume correctness for $t=n$. By definition, $u_{k,c}^{(n+1)}=f_{n+1}(u_{k,c}^{(n)},kv_{k,c}^{(n)})$ where $f_{n+1}$ is a ReLU FNN. By assumption, $v_{k,c}^{(n)}$ is w-describable and so by \Cref{lemma:describe_preserve} we have that $kv_{k,c}^{(n)}$ is describable. Also, by assumption, $u_{k,c}^{(n)}$ is describable. Hence, by \Cref{lemma:describe_preserve} we have that $u_{k,c}^{(n+1)}$ is describable. The proof for $v_{k,c}^{(n+1)}$ is in similar fashion.
\end{proof}

\subsubsection{\Cref{theorem:no_approx_unbounded}}
\textit{
Let $f:\CG_{\Nat^1}\rightarrow\CZ_\Real$ a feature transformation,
such that for every $k,c$ it holds that $f(G_{k,c})(u)=c$. Then, $\text{\sags }\not\approx f$.
}
\begin{proof}
Immediate from combining \Cref{lemma:describable} and \Cref{lemma:one_of_finite}. 
\end{proof}

\subsubsection{\Cref{corola:not_dominant_nat}}
\textit{
Denote by $S$ the set of all multisets over $\PNat$. Let $g:S\rightarrow\Real$ an aggregation such that $\forall a,b\in\PNat\ g({\{a\} \choose b})=a$, that is, $g$ aggregates every homogeneous multiset to its single unique value.
Then, \sags $\nsubsumesp{\Nat}$ g-aggregation GNNs.
}
\begin{proof}
Let $f:\CG_{\Nat^1}\rightarrow\CZ_\Real$ a feature transformation, such that for every featured graph $G$, and for every vertex $v\in V(G)$, it holds that $f(G)(v)\coloneqq g(N(v))$. Then, by \Cref{theorem:no_approx_unbounded}, \sags $\not\approx f$. Clearly, there is a $g$-aggregation GNN that exactly computes $f$.
\end{proof}

Consider another variant of $\{G_{k,c}\}$:
\begin{itemize}
    \item $V(G_{k,c})=\{u_1,\ldots,u_{k^2}\}\cup\{v_1,\ldots,v_k\}$
    \item $E(G_{k,c})=\bigcup_{i\in [k^2], j\in[k]}\{\{u_i,v_j\}\}$
    \item $Z(G_{k,c})=\bigcup_{i\in [k^2]}\{(u_i,0)\}\bigcup_{i\in [k]}\{(v_i,c)\}$
\end{itemize}
Let $\CN$ be an $m$-layer \sagns. We use the notations $u_{k,c}^{(t)}$ and $v_{k,c}^{(t)}$ with similar meaning to before, where $u_{k,c}^{(t)}$ now refers to each of the $u_i$ vertices.
\begin{lemma}\label{lemma:comp_describable}
It holds that $k^2u_{k,c}^{(m)} + kv_{k,c}^{(m)}$
is describable by a set $F$ such that for every $p\in F$ it holds that $p$ does not contain $k^2c$ (with coefficient $\neq 0$).
\end{lemma}
\begin{proof}
We prove the correctness of the following statements for every $t\in[m]$, from which the lemma clearly follows.
\begin{itemize}
    \item [1.] $u_{k,c}^{(t)}$ is describable.
    \item [2.] $v_{k,c}^{(t)}$ is weakly-describable by a set $F$ such that for every $p\in F$ it holds that $p$ does not contain $kc$.
\end{itemize}
Proof is by induction on $t$. Correctness for $t=0$ is clear. Assume correctness for $t=n$.

1. By definition, ${u_{k,c}^{(n+1)}=f_{n+1}(u_{k,c}^{(n)}, kv_{k,c}^{(n)})}$ for some FNN $f_{n+1}$. By the induction assumption, $u_{k,c}^{(n)}$ is describable and clearly $kv_{k,c}^{(n)}$ is also describable. 
Hence, by \Cref{lemma:describe_preserve} we have that $u_{k,c}^{(n+1)}$ is describable.

2. By definition, ${v_{k,c}^{(n+1)}=f_{n+1}(v_{k,c}^{(n)}, k^2u_{k,c}^{(n)})}$ for some FNN $f_{n+1}$. By the induction assumption, $v_{k,c}^{(n)}$ obtains the stated property, and clearly so does $k^2u_{k,c}^{(n)}$. By \Cref{lemma:describe_preserve}, we have that the output of operating $f_{n+1}$ on $v_{k,c}^{(n)}, k^2u_{k,c}^{(n)}$ obtains the stated property.
\end{proof}

\begin{lemma}\label{lemma:gre_no_approx_mro}
Let $f:\CG_{\Nat^1}\rightarrow\Real$ a graph embedding
such that $\forall k,c\ f(G_{k,c})=\frac{kc}{k+1}$.
Let an FNN $\FF$, and define a readout $\Fro\coloneqq f_\FF\circ\mean$.
Then, ${\Fro\circ\text{\sags }\not\approx f}$.
\end{lemma}
\begin{proof}
Let a \sag $\CN$.
By definition, ${\mean\circ\CN(G_{k,c})=\frac{k^2\cdot u_{k,c}^{(m)} + k\cdot v_{k,c}^{(m)}}{k(k+1)}=
\frac{k\cdot u_{k,c}^{(m)} + v_{k,c}^{(m)}}{(k+1)}}$. By \Cref{lemma:comp_describable}, $k\cdot u_{k,c}^{(m)} + v_{k,c}^{(m)}$ is weakly-describable by a set $F'$ such that for every $p\in F'$ it holds that $p$ does not contain $kc$.
Using a similar technique to the one in proof of \Cref{lemma:describe_preserve}, it is not hard to show that 
$f_\FF\circ\mean\circ\CN(G_{k,c})$ is weakly-describable by a set $F$ such that for every $p\in F$ it holds that $p$ does not contain $kc$.
Let any polynomial ${p\in F}$ and let $b\in\Real$ be the coefficient of $k$ in $p$. It is not hard to verify that for every $c$ it holds that ${\lim_{k\rightarrow\infty}\abs{\frac{p(k,c)}{k+1}}\in\{0,|b|,\infty\}}$. The finiteness of $F$ implies that there is a maximal such $|b|$ over all $p\in F$, denote it by $b_{max}$. The finiteness of $F$ also implies that:
\begin{itemize}
    \item [1.] Given $c$ and $\delta>0$ there exists $K_0$ such that for every $l>K_0$ and every $p\in F$ with a finite limit (as ${k\rightarrow\infty}$) it holds that ${\abs{\frac{p(l,c)}{l+1}-\lim_{k\rightarrow\infty}\frac{p(k,c)}{k+1}}<\delta}$.
    \item [2.] Given $c$ and $\delta>0$ there exists $K_0$ such that for every $l>K_0$ and every $p\in F$ with an infinite limit (as ${k\rightarrow\infty}$) it holds that ${\abs{\frac{p(l,c)}{l+1}-c}>\delta}$.
\end{itemize}
Finally, for every $c$ it holds that ${\lim_{k\rightarrow\infty}\frac{kc}{k+1}=c}$. 
Let $\epsilon>0$, then for $c=\ceil{2\epsilon+b_{max}}$ there exists $k$ such that for every $p\in F$ it holds that ${\abs{\frac{p(k,c)-kc}{k+1}}>\epsilon}$.
\end{proof}

\begin{lemma}\label{lemma:gre_no_approx_sro}
Let $f:\CG_{\Nat^1}\rightarrow\Real$ a graph embedding
such that $\forall k,c\ f(G_{k,c})=\frac{kc}{k+1}$.
Let an FNN $\FF$, and define a readout $\Fro\coloneqq f_\FF\circ\lsum$.
Then, ${\Fro\circ\text{\sags }\not\approx f}$.
\end{lemma}

\begin{proof}
Let $\epsilon>0$, then ${\lsum \circ \CN(G_{k,c})=k^2\cdot u_{k,c}^{(m)} + k\cdot v_{k,c}^{(m)}}$. Clearly, $k^2\cdot u_{k,c}^{(m)} + k\cdot v_{k,c}^{(m)}$ is describable. 
Hence, by \Cref{lemma:describe_preserve}, it holds that $f_\FF\circ\lsum\circ\CN(G_{k,c})$ is describable.
Let $F$ a describing set of $k^2\cdot u_{k,c}^{(m)} + k\cdot v_{k,c}^{(m)}$, let any polynomial $p\in F$, and let $b\in\Real$ be the coefficient of $k^0$ in $p$.
It is not hard to verify that for every $c$ it holds that $\lim_{k\rightarrow\infty}\abs{p(k,c)}\in\{0,|b|,\infty\}$. 
The finiteness of $F$ implies that there is a maximal such $|b|$ over all $p\in F$, denote it by $b_{max}$.
The finiteness of $F$ also implies that:
\begin{itemize}
    \item [1.] Given $c$ and $\delta>0$ there exists $K_0$ such that for every $l>K_0$ and every $p\in F$ with a finite limit (as ${k\rightarrow\infty}$) it holds that ${p(l,c)-\lim_{k\rightarrow\infty}p(k,c)<\delta}$.
    \item [2.] Given $c$ and $\delta>0$ there exists $K_0$ such that for every $l>K_0$ and every $p\in F$ with an infinite limit (as ${k\rightarrow\infty}$) it holds that ${\abs{p(l,c)-c}>\delta}$.
\end{itemize}
Finally, for every $c$ it holds that ${\lim_{k\rightarrow\infty}\frac{kc}{k+1}=c}$. 
Let $\epsilon>0$, then for $c=\ceil{2\epsilon+b_{max}}$ there exists $k$ such that for every $p\in F$ it holds that ${\abs{\frac{p(k,c)-kc}{k+1}}>\epsilon}$.
Let $\epsilon>0$, then for $c=\ceil{2\epsilon+b_{max}}$ there exists $k$ such that for every $p\in F$ it holds that ${\abs{p(k,c)- \frac{kc}{k+1}}>\epsilon}$.
\end{proof}

\subsubsection{\Cref{theorem:gre_no_approx}}
\textit{
Let $f:\CG_{\Nat^1}\rightarrow\Real$ a graph embedding
such that $\forall k,c\ f(G_{k,c})=\frac{kc}{k+1}$.
Let an aggregation ${\Fa\in\{\lsum,\mean\}}$ and an FNN $\FF$, and define a readout $\Fro\coloneqq f_\FF\circ \Fa$.
Then, ${\Fro\circ\text{\sags }\not\approx f}$.
}
\begin{proof}
    Follows from combining \Cref{lemma:gre_no_approx_mro} and \Cref{lemma:gre_no_approx_sro}.
\end{proof}

\subsubsection{\Cref{corola:gre_not_dominant_nat}}
\textit{
Denote by $S$ the set of all multisets over $\PNat$. Let $g:S\rightarrow\Real$ an aggregation such that $\forall a,b\in\PNat\ g({\{a\} \choose b})=a$. Let an aggregation ${\Fa\in\{\lsum,\mean\}}$ and an FNN $\FF$, and define a readout $\Fro\coloneqq f_\FF\circ \Fa$. Then, ${\Fro\circ\text{\sags }\nsubsumesp{\Nat} \mean \circ\ g\text{-GNNs}}$.    
}
\begin{proof}
Clearly, for a straightforward g-aggregation GNN $\CN_g$ it holds that $\CN_g(G_{k,c})(u_i)=c$ and $\CN_g(G_{k,c})(v_i)=0$, hence $\mean \circ \CN_g(G_{k,c})=\frac{k^2c}{k^2+k}=\frac{kc}{k+1}$.
By \Cref{theorem:gre_no_approx}, no composition of $\Fro$ with a \sag can approximate $f(G)=\CN_g(G)$.
\end{proof}

\subsection*{\Cref{subsec:simple}}
We define a new series of featured graphs $\{G_{k,c}\}$ (see \Cref{figure:two_level_sv}). For every $(k,c)\in\PNat^2$:
\begin{itemize}
    \item $V(G_{k,c})=\{u\}\cup\{v_1,\ldots,v_k\}\cup\{w_1,\ldots,w_c\}$
    \item $E(G_{k,c})=\bigcup_{i\in [k]}\{\{u,v_i\}\}\bigcup_{i\in[k],j\in [c]}\{\{v_i,w_j\}\}$
    \item $Z(G_{k,c})= \{(u,1)\}\bigcup_{i\in [k]}\{(v_i,1)\}\bigcup_{i\in [c]}\{(w_i,1)\}$
\end{itemize}
Let $\CN$ be an $m$-layer \sagns. We define $u^{(t)}_{k,c}\coloneqq\CN^{(t)}(G_{k,c},u)$, $v^{(t)}_{k,c}\coloneqq\CN^{(t)}(G_{k,c},v_i)$, and $w^{(t)}_{k,c}\coloneqq\CN^{(t)}(G_{k,c},w_i)$, following a reasoning similar to \Cref{subsec:unbounded}, and view $u^{(t)}_{k,c}, v^{(t)}_{k,c}, w^{(t)}_{k,c}$ as functions of $k,c$

\begin{lemma}\label{lemma:describable_two_level}
It holds that $u_{k,c}^{(m)}$ is describable.
\end{lemma}
\begin{proof}
We show by induction that for every $t\in[m]$ it holds that $v_{k,c}^{(t)}$ is w-describable and that $u_{k,c}^{(t)}, w_{k,c}^{(t)}$ are describable. For $t=0$ we have $u_{k,c}^{(t)}=v_{k,c}^{(t)}=w_{k,c}^{(t)}=1$ and the assumption holds. Assume correctness for $t=n$.
By definition, $u_{k,c}^{(n+1)}=f_{n+1}(u_{k,c}^{(n)},kv_{k,c}^{(n)})$ where $f_{n+1}$ is a ReLU FNN. By assumption, $v_{k,c}^{(n)}$ is w-describable and so by \Cref{lemma:describe_preserve} we have that $kv_{k,c}^{(n)}$ is describable. Also by assumption, $u_{k,c}^{(n)}$ is describable. Hence, by \Cref{lemma:describe_preserve} we have that $u_{k,c}^{(n+1)}$ is describable. For $v_{k,c}^{(n+1)}$, by definition, $v_{k,c}^{(n+1)}=f_{n+1}(v_{k,c}^{(n)},cw_{k,c}^{(n)} + u_{k,c}^{(n)})$, and by assumption $u_{k,c}^{(n)},v_{k,c}^{(n)},w_{k,c}^{(n)}$ are w-describable. Hence, by \Cref{lemma:describe_preserve} we have that $v_{k,c}^{(n+1)}$ is w-describable. The proof for $w_{k,c}^{(n+1)}$ is in similar fashion.
\end{proof}

\subsubsection{\Cref{theorem:no_approx_bounded}}
\textit{
Let $f:\CG_{1}\rightarrow\CZ_\Real$ a feature transformation, such that for every $k,c$ it holds that $f(G_{k,c})(u)=c$. Then, \sags $\not\approx f$.
}
\begin{proof}
Immediate from combining \Cref{lemma:describable_two_level} and \Cref{lemma:one_of_finite}. 
\end{proof}

\subsubsection{\Cref{corola:not_dominant_one}}
\textit{
Denote by $S$ the set of all multisets over $\PNat$, and let $g:S\rightarrow\Real$ an aggregation such that $\forall a,b\in\PNat\ g({\{a\} \choose b})=a$. Then, \sags $\nsubsumesp{\{1\}}$ (Sum, g)-GNNs.
}
\begin{proof}
Let $f:\CG_{\{1\}}\rightarrow\CZ_\Real$ a feature transformation such that for every featured graph $G$, for every vertex $v\in V(G)$, it holds that $f(G)(v)\coloneqq g(\{\lsum(w) : w\in N(v)\})$. Then, by \Cref{theorem:no_approx_bounded}, \sags $\not\approx f$.
Clearly, there is a GNN that uses Sum aggregation in its first layer and $g$ aggregation in its second layer, that exactly computes $f$.
\end{proof}

We define one last variant of a $\{G_{k,c}\}$ series:
\begin{itemize}
    \item $V(G_{k,c})=\{u_1,\ldots,u_{k^2}\}\cup\{v_1,\ldots,v_{k^3}\}\cup\{w_1,\ldots,w_{kc}\}$
    \item $E(G_{k,c})=\bigcup_{j\in[k^2],i\in [k^3]}\{\{u_j,v_i\}\}\bigcup_{i\in[k^3],j\in [kc]}\{\{v_i,w_j\}\}$
    \item $Z(G_{k,c})= \bigcup_{i\in[k^2]}\{(u_i,0)\}\bigcup_{i\in [k^3]}\{(v_i,0)\}\bigcup_{i\in [kc]}\{(w_i,1)\}$
\end{itemize}
Let $\CN$ be an $m$-layer \sagns. The notations $u_{k,c}^{(t)}$, $v_{k,c}^{(t)}$, and $w_{k,c}^{(t)}$, are used as before.

\begin{lemma}\label{lemma:another_describable}
It holds that $k^2u_{k,c}^{(m)} + k^3v_{k,c}^{(m)} + kcw_{k,c}^{(m)}$ is describable by a set $F$ and for every $p\in F$ it holds that $p$ does not contain $k^3c$ (with coefficient $\neq 0$).
\end{lemma}
\begin{proof}
We prove the correctness of the following statements, from which the lemma clearly follows.
\begin{itemize}
    \item [1.] $u_{k,c}^{(t)}$ is weakly-describable by a set $F$ such that for every $p\in F$ it holds that $p$ does not contain $kc$.
    \item [2.] $v_{k,c}^{(t)}$ is describable.
    \item [3.] $w_{k,c}^{(t)}$ is weakly-describable by a set $F$ such that for every $p\in F$ it holds that $p$ does not contain $k^2$. 
    
\end{itemize}
Proof is by induction on $t$. Correctness for $t=0$ is immediate. Assume correctness for $t=n$.

1. By definition, ${u_{k,c}^{(n+1)}=f_{n+1}(u_{k,c}^{(n)}, k^3v_{k,c}^{(n)})}$ for some FNN $f_{n+1}$. By the induction assumption, $u_{k,c}^{(n)}$ obtains the stated property and the same holds for $k^3v_{k,c}^{(n)}$. By \Cref{lemma:describe_preserve}, we have that the output of operating $f_{n+1}$ on $u_{k,c}^{(n)}, k^3v_{k,c}^{(n)}$ obtains the stated property.

2. By definition, ${v_{k,c}^{(n+1)}=f_{n+1}(v_{k,c}^{(n)}, k^2u_{k,c}^{(n)} + kcw_{k,c}^{(n)})}$ for some FNN $f_{n+1}$. By the induction assumption, $v_{k,c}^{(n)}$ obtains the stated property, and clearly so do $k^2u_{k,c}^{(n)}, kcw_{k,c}^{(n)}$. The rest follows similarly to the end of (1).

3. By definition, ${w_{k,c}^{(n+1)}=f_{n+1}(w_{k,c}^{(n)}, k^3v_{k,c}^{(n)})}$ for some FNN $f_{n+1}$. By the induction assumption, $w_{k,c}^{(n)}$ obtains the stated property, and clearly so does $k^3v_{k,c}^{(n)}$.  The rest follows similarly to the end of (1).
\end{proof}

\begin{lemma}\label{lemma:gre_no_approx_mro_sv}
Let $f:\CG_{\{0,1\}^1}\rightarrow\Real$ a graph embedding
such that $\forall k,c\ f(G_{k,c})=\frac{(k^2+kc)kc}{k^3+k^2+kc}$.
Let an aggregation ${\Fa\in\{\lsum,\mean\}}$ and an FNN $\FF$, and define a readout $\Fro\coloneqq f_\FF\circ \Fa$.
Then, ${\Fro\circ\text{\sags }\not\approx f}$.
\end{lemma}
\begin{proof}
Let $\epsilon>0$. Define ${A\coloneqq k^2u_{k,c}^{(m)} + k^3v_{k,c}^{(m)} + kcw_{k,c}^{(m)}}$, then ${\mean\circ\CN(G_{k,c})=\frac{A}{k^3+k^2+kc}}$. By \Cref{lemma:another_describable}, $A$ is describable by a set $F'$ such that for every $p\in F'$ it holds that $p$ does not contain $k^3c$, hence $\mean\circ\CN(G_{k,c})$ is describable.
Hence, by \Cref{lemma:describe_preserve} $f_\FF\circ\mean\circ\CN(G_{k,c})$ is describable.
Let $F$ a describing set be . Let any polynomial $p\in F$ and let $b\in\Real$ the coefficient of the component $k^3$ in $p$. Then, it is not hard to verify that for every $c$ it holds that ${\lim_{k\rightarrow\infty}\abs{\frac{p(k,c)}{k^3+k^2+kc}}\in\{0,|b|,\infty\}}$. The finiteness of $F$ implies that there is a maximal such $|b|$ over all ${p\in F}$, denote it by $b_{max}$. 
The finiteness of $F$ also implies that:
\begin{itemize}
    \item [1.] Given $c$ and $\delta>0$ there exists $K_0$ such that for every $l>K_0$ and every $p\in F$ with a finite limit (as ${k\rightarrow\infty}$) it holds that ${\abs{\frac{p(l,c)}{l^3+l^2+lc}-\lim_{k\rightarrow\infty}\frac{p(k,c)}{k^3+k^2+kc}}<\delta}$.
    \item [2.] Given $c$ and $\delta>0$ there exists $K_0$ such that for every $l>K_0$ and every $p\in F$ with an infinite limit (as ${k\rightarrow\infty}$) it holds that ${\frac{p(l,c)}{l^3+l^2+lc}-c>\delta}$.
\end{itemize}
Finally, for every $c$ it holds that ${\lim_{k\rightarrow\infty}\frac{(k^2+kc)kc}{k^3+k^2+kc}=c}$. Hence, for $c=\ceil{2\epsilon+b_{{max}}}$ there exists $k$ such that for every $p\in F$ it holds that ${\abs{\frac{p(k,c)-(k^2+kc)kc}{k^3+k^2+kc}}>\epsilon}$, implying ${\abs{\mean\ \circ\CN(G_{k,c})-f(G_{k,c})}>\epsilon}$.
\end{proof}

\begin{lemma}\label{lemma:gre_no_approx_sro_sv}
Let $f:\CG_{\Nat^1}\rightarrow\Real$ a graph embedding,
such that for every $k,c$ it holds that $f(G_{k,c})=\frac{(k^2+kc)kc}{k^3+k^2+kc}$. Then, $\lsum \circ$ \sags $\not\approx f$.
\end{lemma}
\begin{proof}
Let $\epsilon>0$. Clearly, ${k^2u_{k,c}^{(m)} + k^3v_{k,c}^{(m)} + kcw_{k,c}^{(m)}}$ is describable.
Let $F$ a describing set of ${k^2u_{k,c}^{(m)} + k^3v_{k,c}^{(m)} + kcw_{k,c}^{(m)}}$, let any polynomial $p\in F$, and let $b\in\Real$ be the coefficient of $k^0$ in $p$. Then, it is not hard to verify that for every $c$ it holds that ${\lim_{k\rightarrow\infty}\abs{p(k,c)}\in\{0,|b|,\infty\}}$. The finiteness of $F$ implies that there is a maximal such $|b|$ over all $p\in F$, denote it by $b_{max}$.
The finiteness of $F$ also implies that:
\begin{itemize}
    \item [1.] Given $c$ and $\delta>0$ there exists $K_0$ such that for every $l>K_0$ and every $p\in F$ with a finite limit (as ${k\rightarrow\infty}$) it holds that ${\abs{p(l,c)-\lim_{k\rightarrow\infty}p(k,c)}<\delta}$.
    \item [2.] Given $c$ and $\delta>0$ there exists $K_0$ such that for every $l>K_0$ and every $p\in F$ with an infinite limit (as ${k\rightarrow\infty}$) it holds that ${\abs{p(l,c)-c}>\delta}$.
\end{itemize}
Finally, for every $c$ it holds that ${\lim_{k\rightarrow\infty}\frac{(k^2+kc)kc}{k^3+k^2+kc}=c}$. Hence, for $c=\ceil{2\epsilon+max(0,b_{{max}})}$ there exists $k$ such that for every $p\in F$ it holds that ${\abs{p(k,c)-\frac{(k^2+kc)kc}{k^3+k^2+kc}}>\epsilon}$, implying ${\abs{\lsum\ \circ\CN(G_{k,c})-f(G_{k,c})}>\epsilon}$.
\end{proof}

\subsubsection{\Cref{theorem:gre_no_approx_sv}}
\textit{
Let $f:\CG_{\{0,1\}^1}\rightarrow\Real$ a graph embedding
such that $\forall k,c\ f(G_{k,c})=\frac{(k^2+kc)kc}{k^3+k^2+kc}$.
Let an aggregation ${\Fa\in\{\lsum,\mean\}}$ and an FNN $\FF$, and define a readout $\Fro\coloneqq f_\FF\circ \Fa$.
Then, ${\Fro\circ\text{\sags }\not\approx f}$.
}
\begin{proof}
    Follows from combining \Cref{lemma:gre_no_approx_mro_sv} and \Cref{lemma:gre_no_approx_sro_sv}.
\end{proof}

\subsubsection{\Cref{corola:gre_not_dominant_nat_sv}}
\textit{
Denote by $S$ the set of all multisets over $\PNat$. Let ${g:S\rightarrow\Real}$ an aggregation such that 
${\forall a,b\in\PNat\ g({\{a\} \choose b})=a}$. Let an aggregation ${\Fa\in\{\lsum,\mean\}}$ and an FNN $\FF$, and define a readout $\Fro\coloneqq f_\FF\circ \Fa$. Then, ${\Fro\circ\text{\sags }\nsubsumesp{\{0,1\}} \mean \circ\ \text{(Sum, g)-GNNs}}$.
}
\begin{proof}
Clearly, for a straightforward stereo aggregation (Sum,g)-GNN $\CN_g$ it holds that $\CN_g(G_{k,c})(u_i)=kc$, $\CN_g(G_{k,c})(v_i)=0$, and $\CN_g(G_{k,c})(w_i)=kc$, hence $\mean \circ \CN_g(G_{k,c})=\frac{(k^2+kc)kc}{k^3+k^2+kc}$. 
By \Cref{theorem:gre_no_approx_sv}, no composition of $\Fro$ with a \sag can approximate the graph embedding $f(G)\coloneqq \mean\circ\CN_g(G)$.
\end{proof}

\subsection*{Proofs for \Cref{sec:more_sum}}
\subsubsection{\Cref{lemma:piecewise_polynomial}}
\textit{
Let $\CA$ an $m$-layer \mupa architecture, let $l$ be the maximum depth of any FNN in $\CA$, and let $d$ be the maximum in-degree of any node in any FNN in $\CA$. Then, there exists $r\in\Nat$ such that: for every GNN $\CN$ that realizes $\CA$ it holds that $\CN(G_k,u)$ is piecewise-polynomial (of $k$) with at most $((d+1)^l)^m$ pieces, and each piece is of degree at most $r$.
}
\begin{proof}
Note the following observations:

a. Let $f_1,f_2$ be piecewise polynomial with $p_1,p_2$ pieces, then a linear combination of $f_1,f_2$ has at most $p_1+p_2$ pieces. This can be seen by considering the set of pieces-joint points of $f_1+f_2$, and noticing that it is the union of such points of $f_1$ and such points of $f_2$.
Accordingly, let $f_1,\ldots,f_d$ be piecewise polynomial with at most $p$ pieces each, then a linear combination of $f_1,\ldots,f_d$ has at most $p\cdot d$ pieces.
    
b. Let $f$ be piecewise polynomial with at most $p$ pieces, then $ReLU(f)$ has at most $p+1$ pieces.

c. Let $g$ be an output of a ReLU FNN of depth $l$ with maximal in-degree $d$ for any node, with inputs which are at most $p$-pieces polynomial each. Then, by (a)+(b), $g$ is piecewise-polynomial with $(((pd+1)d+1)d+1)..\leq p\cdot (d+1)^l$ pieces.

d. Let $f(x)$ be piecewise polynomial with at most $p$ pieces, and let $g(x)$ a polynomial, then $g(f(x))$ is piecewise polynomial, with at most $p$ pieces, each of degree at most $deg(f)deg(g)$

e. Let $f(x)$ be piecewise polynomial with at most $p$ pieces, and let $g(y)$ a polynomial, then $g(xf(x))$ is piecewise polynomial, with at most $p$ pieces, each of degree at most $(deg(f)+1)deg(g)$.

Let $\CN$ be a GNN that realizes $\CA$. We define $u^{(t)}_k\coloneqq\CN^{(t)}(G_k,u)$, the feature of $u\in V(G_k)$ after operating the first $t$ layers of $\CN$. Note that $u^{(m)}_k=\CN(G_k,u)$. For every $i,j\in[k]$ there is an automorphism of $G_k$ that maps $v_i$ to $v_j$, thus they receive the same feature throughout the computation. We define $v^{(t)}_k\coloneqq\CN^{(t)}(G_k,v_i)$ for every ${i\in[k]}$. In our argumentation, we view $u^{(t)}_k, v^{(t)}_k$ as functions of $k$.

Using observations [a..e] above, we prove by induction on $t$ that $v_k^{(t)},u_k^{(t)}$, in each coordinate, are piecewise polynomial in $k$ with no more than $((d+1)^l)^t$ pieces, each of degree at most $r_t$ for some $r_t\in\Nat$. For $t=0$ we have that $v_k^{(t)},u_k^{(t)}$ are constants. Assume correctness for $t=n$. By definition, ${u_k^{(n+1)}=f_{n+1}(u_k^{(n)},\Fa^{(n+1)}_1,\ldots,\Fa^{(n+1)}_{b_{n+1}})}$ where $\Fa^{(n+1)}_{j}$ is a shorthand for the aggregation value ${\Fa^{(n+1)}_{j}(\{v_{k,c}^{(n)}\}^{k})}$. By (d),(e), and the induction assumption, each of the input coordinates to $f_{n+1}$ is piecewise polynomial in $k$ with at most $((d+1)^l)^n$ pieces, each of degree at most $r_{n+1}$ for some $r_{n+1}\in\Nat$. Hence, by (c), each coordinate of $u_{k,c}^{(n+1)}$ has at most $((d+1)^l)^n\cdot(d+1)^l=((d+1)^l)^{n+1}$ pieces, each of degree at most $r_{n+1}$. By similar reasoning, $v_k^{(n+1)}$ can be shown to have no more than $((d+1)^l)^{n+1}$ pieces, each of a certain maximal degree.
\end{proof}

\subsubsection{\Cref{theorem:no_converge_no_approx}}
\textit{
Let $f:\CG_{1}\rightarrow\CZ_\Real$ a feature transformation, and define $g(k)\coloneqq f(G_k)(u)$. 
Assume that $g$ does not converge to any polynomial, that is, there exists $\epsilon>0$ such that for every polynomial $p$, for every $K_0$, there exists $k\geq K_0$ such that $\abs{g(k)-p(k)}\geq \epsilon$. Then, \mupasns$\not\approx f$.
}
\begin{proof}
Let an $\epsilon$ by which $g$ does not get forever close to any polynomial, and let a \mupa $\CN$. By \Cref{lemma:piecewise_polynomial}, there is a $K_0$ such that for every $k\geq K_0$ it holds that $\CN(G_k,u)=p(k)$ for some polynomial $p$. By assumption, there exists $k>K_0$ such that $\abs{g(k)-p(k)}\geq \epsilon$. Hence, $\abs{\CN(G_{k},u)-f(G_{k},u)}\geq\epsilon$.
\end{proof}

\subsubsection{\Cref{lemma:polynomial_approximation}}
\textit{
For $x,k\in\Nat$ define $I_{x,k}\coloneqq \{x,x+1,\ldots,x+k-1\}$ the set of consecutive $k$ integers starting at $x$.
Let $f:\mathbb{N}\rightarrow\mathbb{R}$ be a PIL, let $n\in\mathbb{N}$, and define $k_n\coloneqq$
$$1+\max(k:\forall p\in P_n\; \forall x\in\Nat \; \forall y\in[x..(x+k-1)] \; f(y)=p(y))$$
Then, for every $x\in\mathbb{N}$ there exists $\epsilon_{x,k_n}>0$ such that: for every $p\in P_n$ there exists $y\in I_{x,k_n}$ for which $\abs{p(y)-f(y)}\geq\epsilon_{x,k_n}$. That is, for every starting point $x$ there is a bounded interval $I_{x,k_n}$, and a gap $\epsilon_{x,k_n}$, such that no polynomial of degree $\leq n$ can approximate $f$ on that interval below that gap.
}
\begin{proof}
Define $I\coloneqq I_{x,k_n}$. For a real-valued function $h$ whose domain contains $I$, we define $\normi{h}\coloneqq \max(\abs{h(y)} : y\in I_{x,k_n})$, the maximum absolute value $h$ attains on $I_{x,k_n}$.
Define ${\epsilon_{x,k_n}\coloneqq \inf(\normi{f-p} : p\in P_n)}$, the distance of $f$ from the closest  polynomial of degree $\leq n$, in the segment $I_{x,k_n}$. We need to show that $\epsilon_{x,k_n}>0$.
For a vector $a=(a_0,\ldots,a_n)\in\Real^{n+1}$ denote by $\norme{a}$ the Euclidean norm of $a$. For $a,b\in\Real^{n+1}$ we use $d(a,b)\coloneqq\norme{a-b}$ as the metric in our continuity argumentation. Define $p_a(x)\coloneqq a_0+\cdots+a_nx^n$ the polynomial determined by $a$. Note the following:
\begin{itemize}
    \item [a)]For $a\in\Real^{n+1}$, let $g(a)\coloneqq\normi{p_a}$, then $g$ is continuous. 

    \item [b)]For $a\in\Real^{m+1}$, let $g(a)\coloneqq\normi{f-p_a}$, then $g$ is continuous. 
    \item [c)] There exists $T\in\Real$ such that $$\epsilon_{x,k_n}=\inf(\normi{f-p_a} : \norme{a}\leq T)$$
    
    Proof: Let $S=\{a\in\Real^{n+1} : \norme{a}=1\}$ and define ${\delta_S\coloneqq\inf(\normi{p_a} : a\in S)}$. By (a), $\normi{p_a}$ is continuous, and as $S$ is compact we have that there exists $a^*\in S$ such that $\normi{p_{a^*}}=\delta_S$. Note that necessarily $k_n\geq n+1$, then by definition of $\normi{p_{a^*}}$ it must be that either $\delta_S>0$ or $p_{a^*}=0$. Since $a^*\in S$
    , necessarily it is the former that holds. Hence, for every $a\in\Real^{n+1}$ we have that $\normi{p_{a/\norme{a}}}\geq\delta_S$, and by $\normi{p_a}=\norme{a}\cdot\normi{p_{a/\norme{a}}}$ we have $\normi{p_a}\xrightarrow[\norme{a}\rightarrow\infty]{}\infty$. Finally, note that $\normi{f- p_a}\geq \normi{p_a}-\normi{f}$, and let $T$ such that $\norme{a}\geq T \Rightarrow \normi{p_a} > \epsilon_{x,k_n}+1+\normi{f}$, then for all $a : \norme{a}\geq T$ we have $\normi{f- p_a}\geq \epsilon_{x,k_n}+1+\normi{f} - \normi{f}=\epsilon_{x,k_n}+1$. Hence, ${\inf(\normi{f-p} : p\in P_n)=\inf(\normi{f-p_a} : \norme{a}\leq T)}$.
\end{itemize}
By (b) and (c), $\epsilon_{x,k_n}$ is the infimum of a continuous function on a closed ball, hence there exists $a^*\in\Real^{n+1}$ such that $\epsilon_{x,k_n}=\lVert  f-p_{a^*}\rVert_I$. By the assumption that $f$ is PIL, and the definition of $k_n$, we have $\normi{f-p_{a^*}}>0$.
\end{proof}

\input{figures/exp_3D_figure.tex}

\subsubsection{\Cref{lemma:polynomial_approximation_cont}}
\textit{
For every $q,n\in\Nat$ there exists a point $T_{q,n}\in\Nat$ and a gap $\delta_{T_{q,n}}>0$ such that: for every PIL $f:\mathbb{N}\rightarrow\mathbb{R}$, and every piecewise-polynomial $g$ with $q$ many pieces of degree $\leq n$, there exists $y\in\mathbb{N},\; 0\leq y\leq T_{q,n}$ for which $\abs{g(y)-f(y)}\geq\delta_{T_{q,n}}$. That is, the number of pieces and the max degree of a piecewise-polynomial $g$ determine a guaranteed minimum gap by which $g$ misses $f$ within a guaranteed interval.
}
\begin{proof}
Define $T_0=1$. Using the notation of $k_n$ from \Cref{lemma:polynomial_approximation}, for every $i\in[q]$ define \;$T_i\coloneqq (k_n-1)(i)+1$, define $I_i\coloneqq I_{T_{i-1},k_n}$, and define $\delta_i\coloneqq \inf(\lVert{f-p}\rVert I_i : p\in P_n)$. Note that $\delta_i>0$ by \Cref{lemma:polynomial_approximation}. Finally, define ${T_{q,n}\coloneqq T_q}$,\;${\delta_{T_{q,n}}\coloneqq \min(\delta_i:i\in[q])}$. Assume by contradiction that $g$ is close to $f$ by less than $\delta_{T_{q,n}}$ for every $y\in [0..T_{q,n}]$, then, necessarily the first polynomial piece of $g$ ends at most at $T_1-1$, the second at $T_2-1$ and the $q-1$ piece at $T_{q-1}-1$, then the last polynomial piece starts the latest at $T_{q-1}$ and by $T_{q,n}$ it must have missed at least one point by at least $\delta_{T_{q,n}}>0$.
\end{proof}

\subsubsection{\Cref{theorem:no_poly_limited_approx}}
\textit{
Let $f:\CG_{1}\rightarrow\CZ_\Real$ a feature transformation, let $g(k)\coloneqq f(G_k)(u)$, and assume that $g$ is PIL. Then, for every \mupa architecture $\CA$, there exists $\epsilon_{\CA}>0$ such that for every \mupa $\CN$ that realizes $\CA$ there exists $k$ such that $\abs{\CN(G_{k},u)-f(G_{k})(u)}\geq\epsilon$.
}
\begin{proof}
Let the $q,r$ guaranteed by \Cref{lemma:piecewise_polynomial} for $\CA$, and let the $T_{q,r}, \delta_{T_{q,r}}$\; guaranteed by \Cref{lemma:polynomial_approximation_cont} for $q$ pieces of degree $\leq r$. Then, by \Cref{lemma:polynomial_approximation_cont}, for $\epsilon_\CA\coloneqq \delta_{T_{q,r}}$ and $k\coloneqq T_{q,r}$ the statement holds.
\end{proof}

%% file: figures/exp_3D_figure.tex
\begin{figure*}[h!]
\centering
\begin{subfigure}{.47\textwidth}
  \centering
  \includegraphics[width=.99\linewidth]{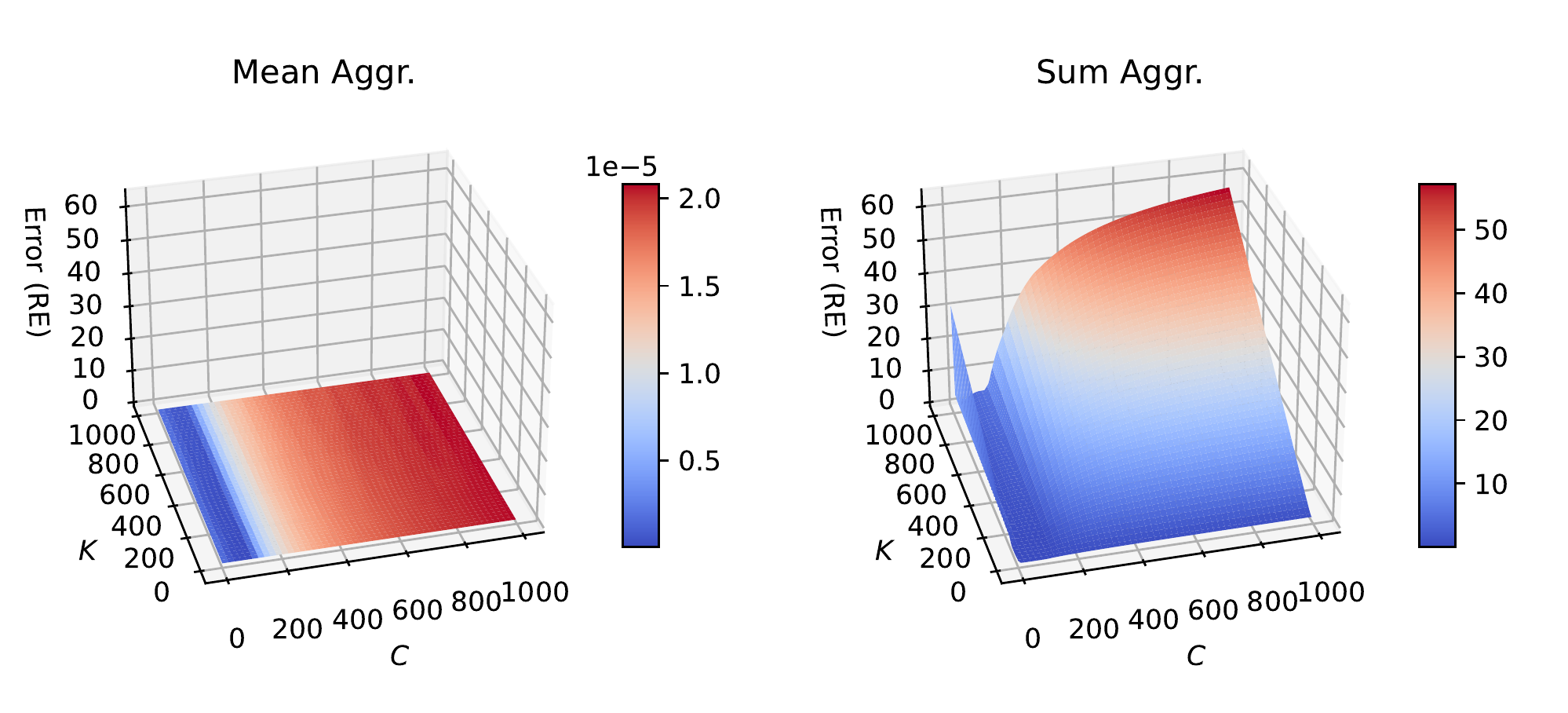}
  \caption{Unbounded Countable Features}
  \label{fig:ucf3D}
\end{subfigure}%
\hfill
\begin{subfigure}{.47\textwidth}
  \centering
  \includegraphics[width=.99\linewidth]{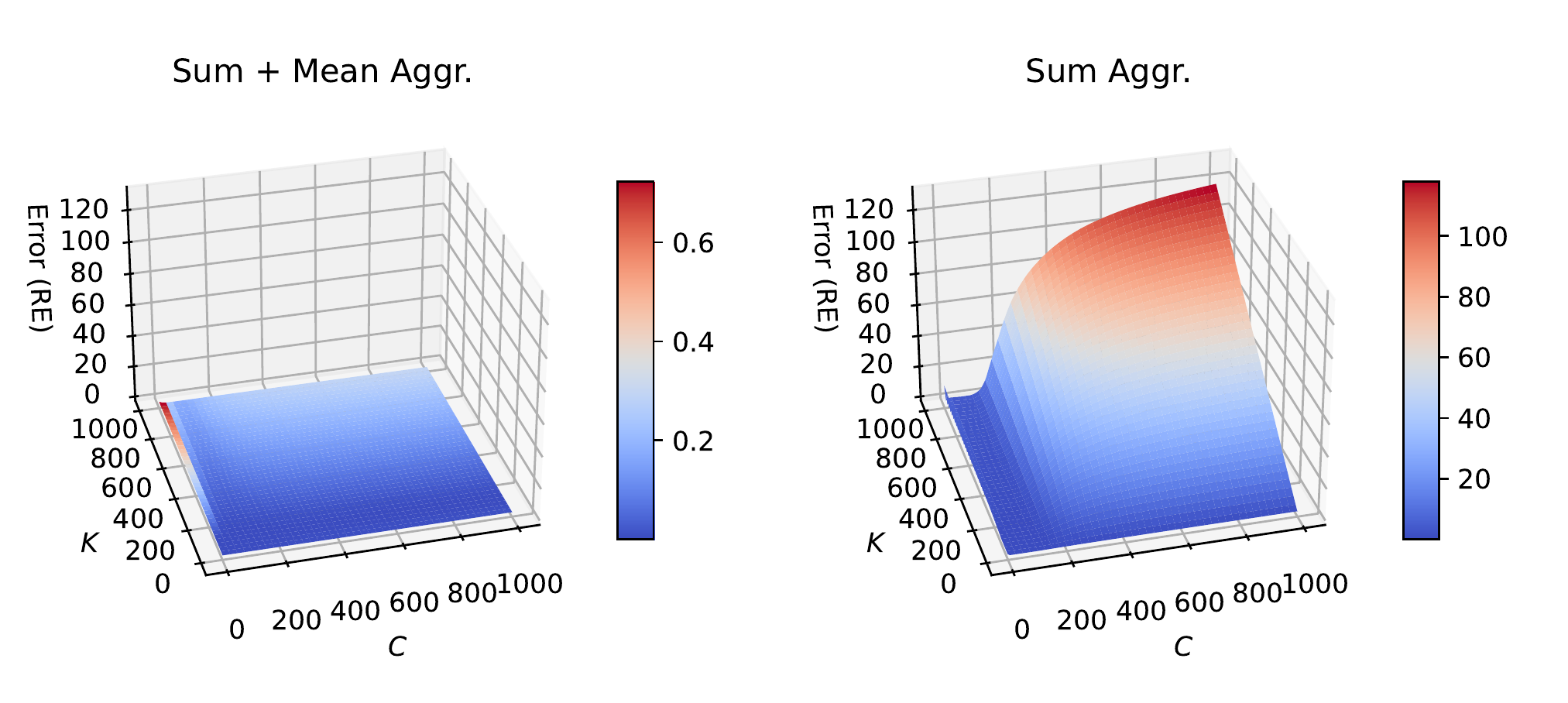}
  \caption{Single Value Features}
  \label{fig:svf3D}
\end{subfigure}
\caption{Relative Error of different aggregations on UC and SV.}
\label{fig:exp3D}
\end{figure*}

%% file: sum_expressivity_experiments_appendix.tex
\section{Experimentation Ext.}
\subsection*{Architecture and Training}
We implement all GNNs using PyTorch Geometric \cite{Fey/Lenssen/2019}.
The update function $f_\FF$ of each GNN layer is a standard 2-layer MLP with a ReLU-activated hidden layer and a linear output layer.
We set the intermediate embedding dimension to 256 and use 2 message passing layers in all models.
We minimize the smooth L1 loss on the training data using the Adam Optimizer \cite{kingma2014adam}.
No readout function is needed.
For both considered graph families the ground truth is a label of the root vertex.
The prediction and loss of all other vertices are simply masked out.

Before each training run we randomly choose 500 graphs from the training data as a validation dataset.
Each model is trained for 500 epochs with a batch size of 100.
The initial learning rate is selected from $\{10^{-3},10^{-4},10^{-5}\}$ based on validation performance.
The learning rate decays with a cosine annealing schedule \cite{loshchilov2016sgdr} throughout training.
We average all results over 5 models trained with different random seeds.
All experiments are conducted on a machine with an NVIDIA RTX A6000 GPU (48GB) and 512GB of RAM running Ubuntu 22.04 LTS. 

\subsection*{Extended Results}
An illustration of the full experimental results can be seen in \cref{fig:exp3D}.
For both datasets, and each tested architecture, we provide the relative error (RE) over the full test range (${k\in [1..1000], c\in [1..1000]}$) as a 3D plot.
The error is provided on the $z$-axis, which is linearly scaled.
The color map is linear as well and is scaled individually for each subplot to highlight additional details.
    
The results for the unbounded countable features (UC) experiment are provided in \cref{fig:ucf3D}.
Note that the color map for the trained \aag is scaled by $10^{-5}$, since the learned function is very close to the ground truth. The trained \sag performs significantly worse.
Relative to itself though, as long as $c$ is in the training range $[1..100]$ it generalizes well along the $k$ axis. Operating the trained \sag, on $c$ in the training range, resembles the bounded input-feature domain setting examined in \Cref{sec:sum_enough}. Hence, the generalization in $k$, when $c$ is in the training range, resembles the result in \Cref{sec:sum_enough}: \sags can approximate Mean when the input-feature domain is bounded. Once $c$ is beyond the training range, the relative error grows rapidly, both along the $k$ axis (for fixed $c$) and along the $c$ axis. Interestingly, the error of the trained \sag also tends upwards at $c<10$. The learned function therefore lacks robustness even towards the lower end of the training range of $c$.

The results for the single value features (SV) experiment are provided in \cref{fig:svf3D}.
Overall, the trained (Sum,Mean)-GNN achieves a significantly lower error than the \sagns.
Like in the UC experiment, as long as $c$ is in the training range $[1..100]$ the trained \sag generalizes relatively well along the $k$ axis, and the performance deteriorates sharply (in both axis) when $c > 100$.
We do note though, that the results of the (Sum,Mean)-GNN in this experiment are substantially worse than those of the \aag in the UC experiment.
While there exists a (Sum,Mean)-GNN that computes exactly the SV-experiment function (see proof of \Cref{corola:not_dominant_one}), Stochastic Gradient Descend (SGD) was not able to learn this function in fine detail. To arrive in a good (Sum,Mean)-GNN instance, the first GNN-layer has to learn to ignore the coordinates of the Mean-aggregation and to use the coordinates of the Sum-aggregation properly, and the second GNN-layer has to learn to ignore the Sum and use the Mean.
These requirements constitute a more challenging learning problem than that of learning a good \aag for the UC task, and the difference is reflected in the results. 
Interestingly, the relative error of the (Sum,Mean)-GNN is worst at the lower end of the training range $c < 10$ for high values of $k$.